\documentclass[english, aps, pre, reprint, notitlepage, superscriptaddress, longbibliography, twocolumn, nofootinbib]{revtex4-2}
\usepackage[T1]{fontenc}
\usepackage[latin9]{inputenc}
\usepackage{amsbsy}
\usepackage{amstext}
\usepackage{amsmath}
\usepackage{amssymb}
\usepackage{graphicx}
\usepackage{esint}
\usepackage{babel}
\usepackage{xcolor}
\usepackage[normalem]{ulem}
\usepackage{array} 
\usepackage[displaymath]{lineno}

\makeatletter
\let\LN@align\align
\let\LN@endalign\endalign
\renewcommand{\align}{\linenomath\LN@align}
\renewcommand{\endalign}{\LN@endalign\endlinenomath}
\let\LN@gather\gather
\let\LN@endgather\endgather
\renewcommand{\gather}{\linenomath\LN@gather}
\renewcommand{\endgather}{\LN@endgather\endlinenomath}
\makeatother

\usepackage[breaklinks=true]{hyperref}
\usepackage{breakcites}
\usepackage{breakurl}

\newcommand{\dd}{\mathrm{d}}

\newcommand{\mat}[1]{\mathbf{#1}}           
\renewcommand{\vec}[1]{\boldsymbol{#1}}     
   
\newcommand{\tsk}{t}                        
\newcommand{\m}{\tsk,\mu}
\newcommand{\n}{\tsk,\nu}

\begin{document}


\title{Dynamics of meta-learning representation in the teacher-student scenario}

\author{Hui Wang}
\affiliation{School of Science, Harbin Institute of Technology (Shenzhen), Shenzhen, 518055, China}

\author{Cho Tung Yip}
\email{h0260416@hit.edu.cn}
\affiliation{School of Science, Harbin Institute of Technology (Shenzhen), Shenzhen, 518055, China}

\author{Bo Li}
\email{libo2021@hit.edu.cn}
\affiliation{School of Science, Harbin Institute of Technology (Shenzhen), Shenzhen, 518055, China}

\begin{abstract}
Gradient-based meta-learning algorithms have gained popularity for their ability to train models on new tasks using limited data. Empirical observations indicate that such algorithms are able to learn a shared representation across tasks, which is regarded as a key factor in their success. However, the in-depth theoretical understanding of the learning dynamics and the origin of the shared representation remains underdeveloped. In this work, we investigate the meta-learning dynamics of nonlinear two-layer neural networks trained on streaming tasks in the teacher-student scenario. Through the lens of statistical physics analysis, we characterize the macroscopic behavior of the meta-training processes, the formation of the shared representation, and the generalization ability of the model on new tasks. The analysis also points to the importance of the choice of certain hyperparameters of the learning algorithms.
\end{abstract}

\maketitle

\section{Introduction}
Supervised learning has long been the dominant paradigm in machine learning and artificial intelligence, yielding substantial achievements across various fields~\cite{Bishop2006}. The primary objective of supervised learning is to obtain a function mapping tailored to a \textit{specific} task by training the machines on a set of input-output examples.  
However, one major limitation of this paradigm is that the trained machine is adapted to the specific task considered and is typically not suited for other tasks, even though they may share some common properties with the existing task. In extreme scenarios, it may require training a new model from scratch for each new task. This is not desirable if the supervised learning tasks require a large number of input-output examples, a common requirement for training large-scale deep neural networks~\cite{LeCun2015, Najafabadi2015}.

To overcome this drawback, many alternative machine learning pipelines have been proposed, such as multitask learning, transfer learning, self-supervised learning, meta-learning, and so on~\cite{Pan2010, Wang2023, Devlin2019, Chen2020, Radford2021, Schmidhuber1987, Bengio1991, Andrychowicz2016, Finn2017, Snell2017}. A common strategy in the recent deep learning practice is to learn a useful representation of the data, which will become applicable to many different downstream tasks~\cite{Devlin2019, Chen2020, Radford2021}. 
Alternatively, meta-learning typically aims at learning to how learn the algorithms for new tasks. Among various meta-learning approaches, the gradient-based meta-learning algorithms have received significant attention~\cite{Andrychowicz2016, Finn2017}. A notable example is the model-agnostic meta-learning (MAML) algorithm~\cite{Finn2017}. It aims to learn a good initialization of the model parameters such that the model can be quickly adapted to new tasks starting from this set of initial model parameters; these initial model parameters are then meta-trained on many different tasks. 
As such, it involves two nested levels of optimization processes, both of which are implemented by gradient descent algorithms in MAML. Subsequent empirical work speculated that the success of MAML is also achieved by learning a shared representation across tasks and further proposed methods to simplify MAML~\cite{Raghu2020Rapid}. 

However, the nested optimization processes make it challenging to develop theoretical analyses for meta-learning algorithms like MAML. To facilitate the analysis, researchers often utilize simplified architectures, such as linear models~\cite{Zou2022, YuHuang2022}. Notably, recent works investigated the learning dynamics of two-layer linear networks and proved that they can successfully learn the hidden representation under the learning dynamics with the population gradients in use~\cite{Collins2022, Yuksel2024}. It is unknown whether similar results can be extended to more realistic settings of nonlinear neural networks and the learning dynamics with finite-sample gradients. 
In this paper, we aim to tackle the challenging scenarios with nonlinear network by employing tools from statistical physics, which has a longstanding tradition in the theoretical development of artificial neural networks~\cite{Watkin1993, Saad1999, Engel2001}, and has regained momentum in recent years~\cite{Bahri2020, Huang2021}. These methods have been recently applied to the multitask learning scenario~\cite{SLee2021, Asanuma2021, LiChan2023, Ingrosso2024, Shan2024} to study phenomena such as catastrophic forgetting and the effects of task similarity, as well as to propose novel learning algorithms; in these studies, the number of tasks is finite and there are abundant examples for each task.
In this work, we focus on a scenario that is more typical in meta-learning, where the number of tasks is large, but the number of examples per task is much smaller than in single-task settings.
We will further consider that the tasks arrive in a streaming, online fashion, which is similar to the usual continual learning setting. However, meta-learning approaches have greater emphasis on the ability to adapt to new tasks and require additional training steps for a new task, which differs from the usual continual learning setting.

To theoretically investigate the above-mentioned meta-learning scenario, we assume that the target functions to be learned for various tasks are generated by a meta-teacher network, and there exists a common latent representation for different tasks. The meta-learner needs to learn the common representation in order to be able to adapt to each specific task based on a few examples. By exploiting the limit of large input dimension and some assumptions of the meta-teacher's parameters, we characterize the learning dynamics of the high dimensional microscopic systems by the evolution of a few macroscopic order parameters, which is much more tractable and allow us to perform detailed theoretical analysis. 
The remaining of the paper is outlined as follows. We introduce the meta-learning problem and algorithm under consideration in Sec.~\ref{sec:meta_learning_setup}, and perform the teacher-student analysis in Sec.~\ref{sec:teacher_student}. We then discuss the results in Sec.~\ref{sec:results} and conclude our work in Sec.~\ref{sec:conclusion}.

\section{The Problem and Algorithm of Meta-learning}\label{sec:meta_learning_setup}
The primary objective of meta-learning is to create a framework where a model can learn to adapt to some new task using a few examples. The model involves a meta-learner which is trained on different tasks in the meta-training phase, aiming to develop a meta-level knowledge that is useful for future tasks. 

In the meta-learning problem setup, there are typically many tasks available, which are denoted as $\{ \mathcal{T}_{\tsk} \}_{\tsk = 1,2,\dots}$ and assumed to be drawn from a certain distribution $\mathcal{P}(\mathcal{T})$. Each task $\mathcal{T}_{\tsk}$ comes with its own training and validation datasets, denoted as $D_{\text{train}}^{\mathcal{T}_{\tsk}}$ and $D_{\text{val}}^{\mathcal{T}_{\tsk}}$, respectively. We focus on supervised learning tasks, so both the training and validation sets consist of input-output example pairs.
Suppose a specific learner $\mathcal{L}_{\tsk}$ is assigned to solve task $\mathcal{T}_{\tsk}$. Its model weights are trained on the training set $D_{\text{train}}^{\mathcal{T}_{\tsk}}$ and evaluated on the validation set $D_{\text{val}}^{\mathcal{T}_{\tsk}}$. If the learner $\mathcal{L}_{\tsk}$ inherits useful meta-level knowledge about the task $\mathcal{T}_{\tsk}$ from an experienced meta-learner, it should require significantly fewer training examples than if it were learning from scratch. The challenge is how to effectively obtain such an experienced meta-learner.
The model-agnostic meta-learning (MAML) algorithm addresses this by maintaining and adapting the meta-learner which serves as a valuable initialization (denoted as $\vec{\theta}_0$) for the model weights $\vec{\theta}_{\tsk}$ of the learner $\mathcal{L}_{\tsk}$.
Since the learner $\mathcal{L}_{\tsk}$ does not start from scratch, but has inherited the good initialization $\vec{\theta}_0$ from the meta-learner, it may require only a few steps of gradient-descent update by using the training set $D_{\text{train}}^{\mathcal{T}_{\tsk}}$ in order to solve the task $\mathcal{T}_{\tsk}$.
The parameters $\vec{\theta}_0$ of the meta-learner are then updated also in a gradient-descent manner in order to reduce the validation error of the tasks that have shown up. Note that the validation error of task $\mathcal{T}_{\tsk}$ is defined on the validation set $D_{\text{val}}^{\mathcal{T}_{\tsk}}$.

Formally, the learner $\mathcal{L}_{\tsk}$ would like to learn a function $f_{\vec{\theta}_{\tsk}}$ parameterized by $\vec{\theta}_{\tsk}$ for solving task $\mathcal{T}_{\tsk}$. It does so by starting from the meta-learner's weights $\vec{\theta}_0$ and performing gradient descent using the training data $D_{\text{train}}^{\mathcal{T}_{\tsk}} = \{(\vec{\xi}^{\m}, \sigma^{\m})\}_{\mu = 1}^{P}$. In MAML, a one-step gradient descent is applied~\cite{Finn2017},
\begin{align}
    \vec{\theta}_{\tsk} = \vec{\theta}_0 - \eta_{\tsk} \big[ \boldsymbol{\nabla}_{ \vec{\theta} }  \ell( \vec{\theta} \mid D_{\text{train}}^{\mathcal{T}_{\tsk}} ) \big] \big\vert_{\vec{\theta} = \vec{\theta}_0}, \label{eq:MAML_inner}
\end{align}
where $\ell(\cdot)$ is the loss function and $\eta_{\tsk}$ is the learning rate for training task $\mathcal{T}_{\tsk}$. For regression tasks, the mean square error (MSE) loss is often used
\begin{align}
    \ell( \vec{\theta} \mid D_{\text{train}}^{\mathcal{T}_{\tsk}} ) = \frac{1}{2 P} \sum_{\mu = 1}^P \big[ f_{\vec{\theta}} (\vec{\xi}^{\m}) - \sigma^{\m} \big]^2,
\end{align}
where $f_{\vec{\theta}} (\cdot)$ is a learner function parameterized by $\vec{\theta}$.
 
The quality of $\vec{\theta}_{\tsk}$ for solving task $\mathcal{T}_{\tsk}$ can be evaluated by inspecting the validation loss $\ell(\vec{\theta}_{\tsk} \mid D_{\text{val}}^{\mathcal{T}_{\tsk}})$ using $D_{\text{val}}^{\mathcal{T}_{\tsk}} = \{(\vec{\xi}^{\n}, \sigma^{\n})\}_{\nu = 1}^{V}$. For regression tasks, we can again use the MSE loss
\begin{align}
    \ell( \vec{\theta}_{\tsk} \mid D_{\text{val}}^{\mathcal{T}_{\tsk}} ) = \frac{1}{2 V} \sum_{\nu = 1}^{ V } \big[ f_{\vec{\theta}_{\tsk}} (\vec{\xi}^{\n}) - \sigma^{\n} \big]^2.\label{eq:outer_loss}
\end{align}
Note that $\vec{\theta}_{\tsk}$ is a function of $\vec{\theta}_0$ according to Eq.~(\ref{eq:MAML_inner}), which can be denoted as $\vec{\theta}_{\tsk}( \vec{\theta}_0 )$. The meta-learner would like to update its parameter $\vec{\theta}_0$ so as to reduce the validation losses of different tasks,
\begin{align}
    \vec{\theta}_0^{\text{next}} & \leftarrow \vec{\theta}_0 - \eta_0 \nabla_{ \vec{\theta}_0 } \sum_{\mathcal{T}_{\tsk}} \ell \big( \vec{\theta}_{\tsk} (\vec{\theta}_0) \mid D_{\text{val}}^{\mathcal{T}_{\tsk}} \big), \label{eq:MAML_outer} 
\end{align}
where $\eta_0$ is the learning rate for meta-training.
Note that the update rule in Eq.~(\ref{eq:MAML_outer}) is designed to minimize the validation loss defined by the validation data sets $\{ D_{\text{val}}^{\mathcal{T}_{\tsk}} \}$. These validation datasets are included because the primary objective of the meta-learner is to guide task-specific learners to effectively \textit{generalize}. Relying solely on the training data sets $\{ D_{\text{train}}^{\mathcal{T}{\tsk}} \}$ would not provide a reliable estimate of generalization ability.

After many episodes of meta-training using Eq.~(\ref{eq:MAML_outer}), we hope to obtain a meta-learner which has pooled the knowledge of different tasks into the meta-learner's parameters $\vec{\theta}_0$. The applicability of this meta-learner should be evaluated on an unseen task $\mathcal{T}_*$ which is assumed be drawn from $\mathcal{P}( \mathcal{T} )$. Its  performance can be assessed by the meta-generalization error $\epsilon^{\text{meta}}_g$, defined as
\begin{align}
    \epsilon^{\text{meta}}_g & = \big\langle \ell\big( \vec{\theta}_* (\vec{\theta}_0) \mid D_{\text{test}}^{\mathcal{T}_*} \big)  \big\rangle_{ \mathcal{T}_* \sim \mathcal{P}( \mathcal{T} ) }, \\
    \vec{\theta}_* ( \vec{\theta}_0 ) & := \vec{\theta}_0 - \eta_* \big[ \boldsymbol{\nabla}_{ \vec{\theta} }  \ell( \vec{\theta} \mid D_{\text{train}}^{\mathcal{T}_*} ) \big] \big\vert_{\vec{\theta} = \vec{\theta}_0},
\end{align}
where $D_{\text{test}}^{\mathcal{T}_*}$ is a test dataset corresponding to the task~$\mathcal{T}_*$. 
As such, the meta-generalization error $\epsilon^{\text{meta}}_g$ measures how good the meta-parameters $\vec{\theta}_0$ are in guiding the parameters $\vec{\theta}_*$ for solving a particular task $\mathcal{T}_*$ drawn from $\mathcal{P}( \mathcal{T} )$.

The MAML algorithm outlined above constitutes a bilevel optimization process, where Eq.~(\ref{eq:MAML_inner}) and Eq.~(\ref{eq:MAML_outer}) are referred to as the inner-loop update and the outer-loop update, respectively. 
In practice, the one-step inner-loop update in Eq.~(\ref{eq:MAML_inner}) has been generalized for better performance~\cite{ZhenguoLi2017, Rajeswaran2019}, but the resulting algorithms are even more convoluted. 
While the MAML algorithm is conceptually simple, the bilevel optimization structure in the original MAML setting has already involved second-order derivatives (e.g., in the outer-loop update), which presents challenges to computations in high dimensions. 
Approximations using only the first-order derivatives have been propose to reduce computations~\cite{Nichol2018}. 
Another study observed that when using deep neural networks for solving computer vision tasks, the meta-learner tends to reuse the features for solving a new task, i.e., without much further tuning the representation~\cite{Raghu2020Rapid}. Based on this empirical observation, the authors  proposed the almost-no-inner-loop (ANIL) algorithm, where only the parameters of the last layer of the neural networks are updated in the inner-loop optimization. 
In light of these earlier works, we focus on the first-order ANIL (FO-ANIL) algorithm for simplification, where the algorithmic implementations will be described below.

\section{Teacher-student Analysis}\label{sec:teacher_student}

In this work, we extend the traditional teacher-student framework for supervised learning to the context of meta-learning. In the study of supervised learning, the core idea of the teacher-student analysis involves designing nontrivial teacher functions as the ground truths. The teacher functions generate input-output examples which are then presented to the student models for learning. It should be noted that the student models do not have access to the internal details of the teacher models, but only to the examples provided by them. In certain cases, the generalization ability of the student model can be expressed analytically through macroscopic order parameters that link the weights of the teacher and student models. This analytical approach enables us to conduct detailed theoretical examinations and gain deeper insights into the learning process~\cite{Engel2001, Saad1999, Huang2021}.

\subsection{The meta-teacher and task-specific teachers}\label{sec:teacher}
To address meta-learning across various tasks, we present a meta-teacher model designed to generate multiple specific teachers tailored to different tasks. These specific teachers share similar representations with the meta-teacher but differ from one another based on the supervised learning tasks built on top of the representation.
Specifically, both the meta-teacher and the specific teachers are two-layer neural networks, having $N$ input units, $M$ hidden units and one output unit. The meta-teacher has input-to-hidden weights $\mat{B} \in \mathbb{R}^{M \times N}$ and hidden-to-output weights $\vec{u}^0 \in \mathbb{R}^{M}$.
In this study, we set $\vec{u}^0$ to zero, indicating that the meta-teacher does not impose a preferred function but instead provides a shared representation mapping.

\begin{figure}[!htbp]
    \centering
    \includegraphics[scale=0.32]{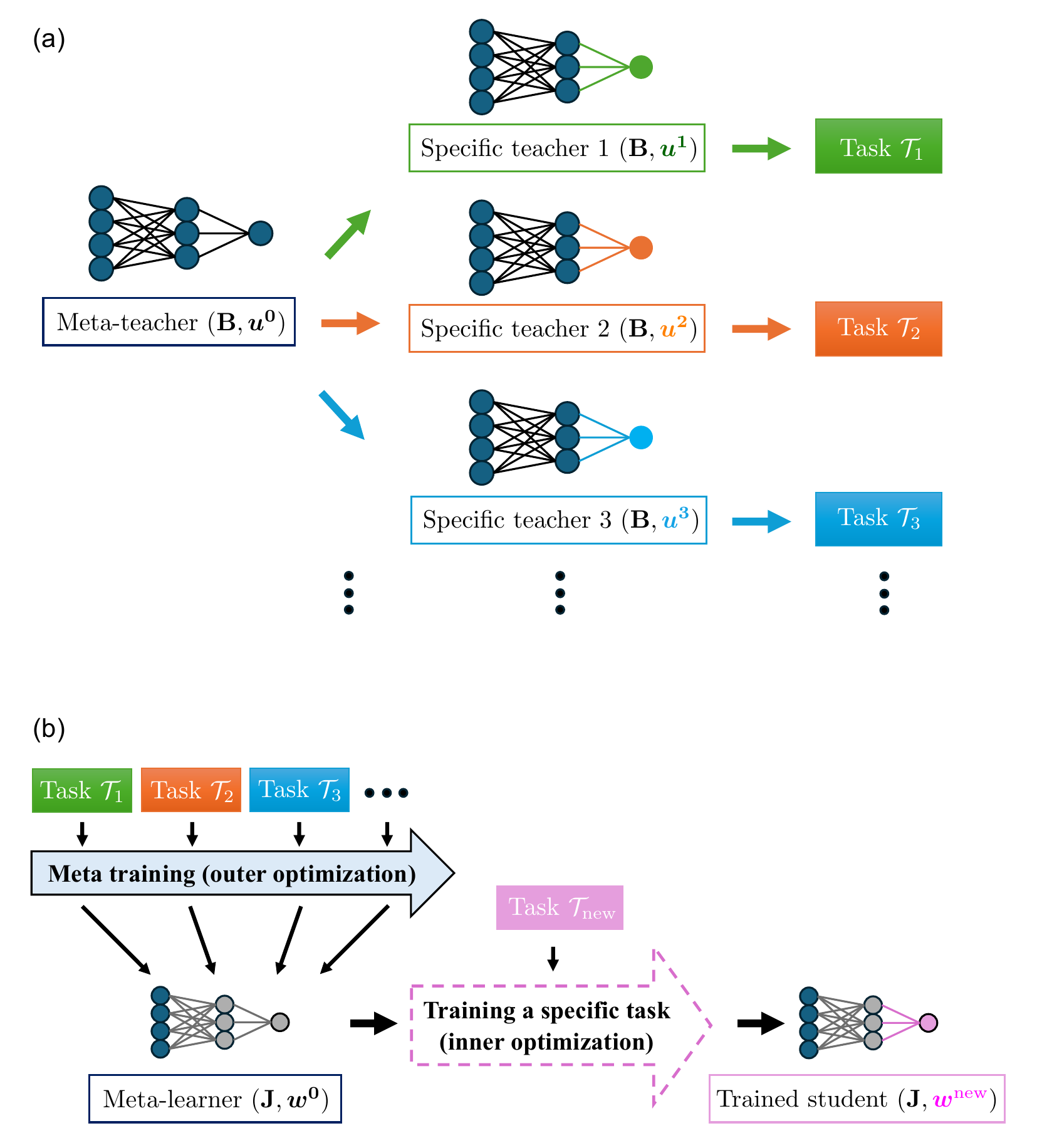} 
    \caption{The framework of the teacher-student analysis for meta-learning under consideration. All models are two-layer neural networks. (a) The meta-teacher generates different task-specific teachers, which share the same input-to-hidden weights with the meta-teacher (denoted as $\mat{B}$) but have different hidden-to-output weights (denoted as $\vec{u}^{\tsk}$). (b) The meta-training and testing processes. The meta-learner is meta-trained on different tasks using the FO-ANIL algorithm. Each specific learner inherits the input-to-hidden weights $\mat{J}$ from the meta-learner, but maintains its own hidden-to-output weights (denoted as $\vec{w}^{\tsk}$) which are task specific. The meta-testing is evaluated by its ability to guide a new learner to solve an unseen task $\mathcal{T}_{\text{new}}$ (through an inner-loop optimization).}
    \label{fig:illustrate}
\end{figure}

For the task-specific teachers, we will first assume that their input-to-hidden weights are the same and copied from the meta-teacher, having values $\mat{B}$. We will relax this constraint in a later section. 
Conversely, the hidden-to-output weights $\vec{u}^{\tsk} \in \mathbb{R}^{M}$ are specific to the $\tsk$th task, assumed to be drawn from a multivariate Gaussian distribution $\vec{u}^{\tsk} \sim \mathcal{N}(0, I_M)$.
In this way, the weights $\{ \vec{u}^{\tsk} \}$ are completely independent among different tasks, which in fact creates a challenging scenario for the meta-learner. A less challenging scenario for the meta-learner is to create some level of similarity among different $\{ \vec{u}^{\tsk} \}$, but we will primarily focus on the regime of uncorrelated $\{ \vec{u}^{\tsk} \}$ to simplify the picture. The teacher-task generation process is illustrated in Fig~\ref{fig:illustrate}(a). In the notation of Sec.~\ref{sec:meta_learning_setup}, task $\mathcal{T}_{\tsk}$ is fixed by the parameters $\{ \mat{B},\vec{u}^{\tsk} \}$ of the $\tsk$th teacher network.

The training dataset for the $\tsk$th task can be prepared by separately feeding $P$ inputs $\{ \vec{\xi}^{\m} \}_{\mu=1}^{P}$ to the $\tsk$th specific teacher network and collecting the outputs
\begin{align}
    \sigma^{\m} & = \frac{1}{\sqrt{M}} \sum_{m=1}^M u^{\tsk}_{m} \; g\big( \vec{B}_{m}^{\top} \cdot \vec{\xi}^{\m} \big), \quad \mu = 1, 2, \dots, P, \label{eq:sigma_t_mu_by_teacher}
\end{align}
where $\vec{B}_{m} \in \mathbb{R}^{1 \times N}$ is the $m$th row of the matrix $\mat{B}$ (or the weight vector of the $m$th hidden unit), and $g(\cdot)$ is the activation function. 
The factor $1/\sqrt{M}$ in Eq.~(\ref{eq:sigma_t_mu_by_teacher}) was introduced to ensure that the outputs remain of the same order of magnitude across networks with different hidden layer widths. The validation or test dataset can be obtained similarly.
We primarily consider the error activation function $g(x) = \operatorname{erf}(x / \sqrt{2})$ for analytical tractability, and take the high-dimension limit of the input $N \to \infty$. Moreover, we concern about only a finite number of hidden units in all the teacher networks [i.e., $M \sim \mathit{O} (1)$]. Another interesting case involves networks of infinite width with $M \to \infty$ (as in~\cite{BL2018, BL2020PRL}), but analyzing the learning dynamics in such cases requires different assumptions and theoretical tools~\cite{Jacot2018, Mei2018, Yang2021}, which are beyond the scope of this study.

For all tasks, the inputs are drawn independently and identically from a spherical Gaussian distribution, i.e., $\vec{\xi} \sim \mathcal{N}(0, I_N)$. While it is possible to extend the uncorrelated Gaussian input distribution to a structured input distribution (as in~\cite{Goldt2020}), doing so would require additional techniques, which are also better suited for future studies.

\subsection{The meta-learner and task-specific students}\label{sec:student}
The meta-learner and the student networks for specific tasks are also assumed to be two-layer neural networks, having $N$ input units, $K$ hidden units and one output unit. We also consider that $K$ is finite [i.e., $K \sim \mathit{O}(1)$], which is not necessarily the same as $M$.
The meta-learner has input-to-hidden weights $\mat{J} \in \mathbb{R}^{K \times N}$ and hidden-to-output weights $\vec{w}^0 \in \mathbb{R}^{1 \times K}$.
For the $\tsk$th task, the student network copies the input-to-hidden weights $\mat{J}$ from the meta-learner, saving the effort to learn the representation mapping, but it maintains its own hidden-to-output weights $\vec{w}^{\tsk} \in \mathbb{R}^{1 \times K}$. Such a setup is in alignment with the ANIL algorithm, where the representation mapping is not fine-tuned when the $\tsk$-student is adapted to solve the $\tsk$-task.

Feeding the inputs $\{ \vec{\xi}^{\m} \}_{\mu=1}^{P}$ to the $\tsk$th student network gives rise to its output predictions
\begin{align}
    \tilde{\sigma}^{\m} = \frac{1}{\sqrt{K}} \sum_{k=1}^K w^{\tsk}_{k} \; g\big( \vec{J}_{k}^{\top} \cdot \vec{\xi}^{\m} \big), \quad \mu = 1, 2, \dots, P.
\end{align}
The objective of the $\tsk$th student is to adapt the weights $\vec{w}^{\tsk}$ such that the above outputs $\{ \tilde{\sigma}^{\m} \}_{\mu=1}^{P}$ get close to the labels $\{ \sigma^{\m} \}_{\mu=1}^{P}$ given in Eq.~(\ref{eq:sigma_t_mu_by_teacher}) in order to solve the task $\mathcal{T}_{\tsk}$.

\subsection{Meta-training and meta-testing}\label{sec:training}
Referring to the notations in Sec.~\ref{sec:meta_learning_setup}, we identify that the model parameters of the $\tsk$th student are $\vec{\theta}_{\tsk} = \{ \mat{J}, \vec{w}^{\tsk} \}$. In the MAML algorithm, both $\mat{J}$ and $\vec{w}^0$ are adapted in the inner-loop optimization when solving the specific task $\mathcal{T}_{\tsk}$. 
On the other hand, in the ANIL algorithm, the input-to-hidden weights $\mat{J}$ which map the inputs to the latent representations are fixed in the inner-loop optimization, as we have exploited in Sec.~\ref{sec:student} for the setup of teacher-student analysis.
Furthermore, the FO-ANIL algorithm neglects any possible second derivatives in the vanilla ANIL algorithm to simplify the computations.

We consider the case that the tasks $\{ \mathcal{T}_1, \mathcal{T}_2, \dots, \mathcal{T}_{\tsk}, \dots \}$ drawn from the distribution $\mathcal{P}(\mathcal{T})$ arrive in a streaming, online fashion where $\tsk$ can be viewed as a time index. Each task comes with its own training and validating data sets, generated by the meta-teacher framework described above using independent inputs $\vec{\xi}^{\m}$. 
For the task $\mathcal{T}_{\tsk}$, the $\tsk$th student network receives a small training set $D_{\text{train}}^{\mathcal{T}_{\tsk}} = \{(\vec{\xi}^{\m}, \sigma^{\m})\}_{\mu=1}^{P}$. The training loss for the inner-loop optimization is defined as
\begin{align}
    & \ell(\vec{w} \mid \mat{J}, D_{\text{train}}^{\mathcal{T}_{\tsk}})  \\
    & = \frac{1}{2P}\sum_{\mu=1}^P \bigg[ \frac{1}{\sqrt{K}} \sum_{k=1}^K w_{k} \; g\big( \vec{J}_{k}^{\top} \cdot \vec{\xi}^{\m} \big) - \sigma^{\m} \bigg]^{2}, \nonumber
\end{align} 
where the prefactor $\frac{1}{2}$ is introduced for notational convenience.
The FO-ANIL algorithm dictates that the $\tsk$th student network slightly adapts the hidden-to-output weights by lowering this training loss starting from the parameters of the meta-learner
\begin{align}
    \vec{w}^{\tsk} = \vec{w}^{0} - \eta_{w}\mathit{\nabla}_{\vec{w}} \big[ \ell(\vec{w} \mid \mat{J}, D_{\text{train}}^{\mathcal{T}_{\tsk}}) \big] \big\vert_{\vec{w} = \vec{w}^{0}}, \label{eq:w_t_ANIL}
\end{align}
where $\eta_w$ is the learning rate for updating the parameters $\vec{w}^{\tsk}$.

We further consider the special case with $\vec{w}^0 = 0$, such that $\vec{w}^{\tsk}$ are mostly driven by the data $D_{\text{train}}^{\mathcal{T}_{\tsk}}$ rather than the ``prior'' knowledge of $\vec{w}^0$ from the meta-learner. In Eq.~(\ref{eq:w_t_ANIL}), we also fix the input-to-hidden weights $\mat{J}$ to be their most updated values $\mat{J}^{\tsk - 1}$, which gives rise to
\begin{align}
    w_{k}^{\tsk} & = \frac{\eta_{w}}{P} \sum_{\mu=1}^P \sigma^{\m} \frac{ g\big( ( \vec{J}_{k}^{t-1} )^{\top} \cdot \vec{\xi}^{\m} \big) }{\sqrt{K}}.
    \label{eq:w_k_of_J}
\end{align}
In this special case, the meta-learner has no need to maintain the parameters $\vec{w}^0$, and the meta-training resorts to learning a better representation mapping $\mat{J}$. To achieve this, we first define the validation loss for the outer optimization using the validation dataset $D_{\text{val}}^{\mathcal{T}_{\tsk}} = \{(\vec{\xi}^{\n}, \sigma^{\n})\}_{\nu = 1}^{V}$
\begin{align}
    & \ell(\mat{J} \mid D_{\text{val}}^{\mathcal{T}_{\tsk}})  \\
    & = \frac{1}{2V}\sum_{\nu=1}^V \bigg[ \frac{1}{\sqrt{K}} \sum_{k=1}^K w_{k}^{\tsk} \; g\big( \vec{J}_{k}^{\top} \cdot \vec{\xi}^{\n} \big) - \sigma^{\n} \bigg]^{2}. \nonumber
\end{align} 
The input-to-hidden layer weights $\mat{J}$ are then updated in order to reduce the validation loss as
\begin{align}
    \vec{J}_{k}^{\tsk} = \vec{J}_{k}^{\tsk - 1} - \frac{ \eta_J }{ N } \big[ \mathit{\nabla}_{\vec{J}_{k}} \ell(\mat{J} \mid D_{\text{val}}^{\mathcal{T}_{\tsk}}) \big] \big\vert_{\vec{J}_k = \vec{J}_k^{\tsk - 1}}, \label{eq:J_outer_update}
\end{align}
where $\eta_J$ is the scaled learning rate for updating the parameters $\vec{J}_k$. As such, the meta-representation mapping $\mat{J}$ is continuously evolving as tasks arrive.

Suppose that at a certain time, the meta-learner obtains a meta-representation mapping $\mat{J}$, where we would like to evaluate its performance. This can be done by first considering a new task $\mathcal{T}_{\text{new}}$ drawn from $\mathcal{P}( \mathcal{T} )$, which comes with a small training set $D_{\text{train}}^{\mathcal{T}_{\text{new}}} = \{(\vec{\xi}^{{\text{new}, \mu}}, \sigma^{{\text{new}, \mu}})\}_{\mu=1}^{P}$ generated by the new task's teacher network. A new student network $\mathcal{L}_{\text{new}}$ will be assigned for this task; it will adapt its hidden-to-output weights according to Eq.~(\ref{eq:w_k_of_J}), yielding $\vec{w}^{\text{new}}$. 
The performance of this new student is measured by the generalization error when predicting an unseen test data point $(\vec{\xi}^{\text{new}, 0}, \sigma^{\text{new}, 0})$, average over the input distribution of $\vec{\xi}^{\text{new}, 0}$
\begin{align}
    & \epsilon_{g}^{\text{new}}(\vec{w}^{\text{new}}, \mat{J} \mid  D_{\text{train}}^{\mathcal{T}_{\text{new}}}) \\ 
    & = \bigg\langle \frac{1}{2} \bigg[ \frac{1}{\sqrt{K}} \sum_{k=1}^K  w^{\text{new}}_{k} g\big( \vec{J}_{k}^{\top} \cdot \vec{\xi}^{\text{new}, 0} \big) - \sigma^{\text{new}, 0} \bigg]^{2} \bigg\rangle_{ \vec{\xi}^{\text{new}, 0} } . \nonumber
\end{align}
The performance of the meta-learner can then be evaluated by the meta-generalization error, which averages over all possible choices of $D_{\text{train}}^{\mathcal{T}_{\text{new}}}$ and $\mathcal{T}_{\text{new}}$
\begin{align}
    \!\!  \epsilon_{g}^{\text{meta}} = \big\langle \big\langle \epsilon_{g}^{\text{new}}(\vec{w}^{\text{new}}, \mat{J} \mid  D_{\text{train}}^{\mathcal{T}_{\text{new}}}) \big\rangle_{ D_{\text{train}}^{\mathcal{T}_{\text{new}}} } \big\rangle_{ \mathcal{T}_{\text{new}} \sim \mathcal{P}( \mathcal{T} ) } .
\end{align}

\subsection{Macroscopic order parameters}
As described above, we have considered the online arrival of task $\mathcal{T}_{\tsk}$ and the independence among all input examples $\vec{\xi}^{\m}$ within and across tasks. The advantage for these considerations is that the meta-learner's weights $\mat{J}^{\tsk - 1}$ up to time $\tsk - 1$ have no statistical dependence on the input examples $\{ \vec{\xi}^{\m} \}$ arrived at time $\tsk$; it makes the analysis significantly easier, which has been exploited in the traditional single-task online learning dynamics~\cite{Biehl1995, SaadPRL1995, Saad1999, Engel2001}.

We denote the preactivation of $k$th hidden unit in the meta-learner's network as $x_k^{\m} = \vec{J}_k^{\top} \cdot \vec{\xi}^{\m}$, and the preactivation of the $m$th hidden unit in the meta-teacher's network as $y_k^{\m} = \vec{B}_k^{\top} \cdot \vec{\xi}^{\m}$. Recall that the elements of $\vec{\xi}^{\m}$ are independent Gaussian random variables with zero mean and unit variance. In the limit where the input dimension is large, i.e., $N \to \infty$, the central limit theorem dictates that the meta-learner's and the meta-teacher's preactivations follow a joint Gaussian distribution, where the means are $\langle x_k^{\m} \rangle = \langle y_m^{\m} \rangle = 0$ and the covariance matrix is specified by 
\begin{align}
    Q_{kl} & := \langle x_k^{\m} x_l^{\m} \rangle = \vec{J}_k^{\top} \cdot \vec{J}_l,  \\
    T_{mn} & := \langle y_m^{\m} y_n^{\m} \rangle = \vec{B}_m^{\top} \cdot \vec{B}_n, \\
    R_{km} &:= \langle x_k^{\m} y_m^{\m} \rangle = \vec{J}_k^{\top} \cdot \vec{B}_m.
\end{align}
More explicitly, the local fields $(\vec{x}^{\m}, \vec{y}^{\m}) \in \mathbb{R}^{K+M}$ obey the following Gaussian distribution
\begin{align}
    \!\!\! \mathcal{P}( \vec{x}^{\m}, \vec{y}^{\m} ) & = \frac{ \exp \left[ -\frac{1}{2} (\vec{x}^{\m}, \vec{y}^{\m})^\top C^{-1} (\vec{x}^{\m}, \vec{y}^{\m}) \right] }{\sqrt{ (2\pi)^{K+M} |C| } }, \\
    C & := \left[ \begin{array}{cc}
        Q & R \\
        R^{\top} & T 
    \end{array} \right],
\end{align}
where the randomness comes from the specific input $\vec{\xi}^{\m}$.

We recall that the label $\sigma^{\m}$ is generated by the $\tsk$th teacher network in Eq.~(\ref{eq:sigma_t_mu_by_teacher}), so the updated hidden-to-output weights $w^{\tsk}_{k}$ of the $\tsk$th student network admit the following expression:
\begin{align}
    w_{k}^{\tsk}( \vec{u}^{\tsk} ) = \frac{\eta_{w}}{P}\sum_{\mu=1}^P\sum_{m=1}^{M}\frac{u_{m}^{\tsk}g(y_{m}^{\m})g(x_{k}^{\m})}{\sqrt{MK}}, \label{eq:w_k_of_xy}
\end{align}
where the randomness comes from the training inputs $\{ \vec{\xi}^{\m} \}_{\mu=1}^{P}$ and the task vector $\vec{u}^{\tsk}$ for task $\mathcal{T}_{\tsk}$.
We note that the dynamics of $w^{\tsk}_k$ are not maintained, as $w^{\tsk}_k$ is not derived from $w^{\tsk-1}_k$. This is in stark contrast to the behavior of hidden-to-output weights in the single-task learning scenario analyzed in Ref.~\cite{Riegler1995}.

Similarly, the updated input-to-hidden weights $\mat{J}$ for the meta-learner in Eq.~(\ref{eq:J_outer_update}) can be expressed as
 \begin{align}
    \vec{J}_{k}^{\tsk} & = \vec{J}_{k}^{\tsk - 1} + \frac{\eta_{J}}{NV}\sum_{\nu=1}^V\frac{h_{k}^{\n}}{\sqrt{K}}  \vec{\xi}^{\n}, \label{eq:J_k_of_delta_k} \\
    h^{\n}_{k} & := \bigg[ \sum_{m=1}^{M}\frac{u_{m}^{\tsk}g(y_{m}^{\n})}{\sqrt{M}}-\sum_{i=1}^{K}\frac{w_{i}^{\tsk}g(x_{i}^{\n})}{\sqrt{K}} \bigg] \nonumber \\
    & \quad \;\; \times w_{k}^{\tsk}g^{\prime}(x_{k}^{\n}), \label{eq:delta_k_of_xy}
\end{align}
where the randomness of the preactivation $\{ x_k^{\n}, y_m^{\n} \}$ comes from the validation inputs $\{ \vec{\xi}^{\n} \}_{\nu=1}^{V}$. 

Since we produce the training inputs $\{ \vec{\xi}^{\m} \}_{\mu=1}^{P}$ and the validation inputs $\{ \vec{\xi}^{\n} \}_{\nu=1}^{V}$ independently, the local fields $( \vec{x}^{\m}, \vec{y}^{\m} )$ in Eq.~(\ref{eq:w_k_of_xy}) and the local fields $( \vec{x}^{\n}, \vec{y}^{\n} )$ in Eq.~(\ref{eq:delta_k_of_xy}) are independent random vectors with the same covariance structure $C$, and their disordered averages can be performed separately.

The evolution of the representation mapping in Eq.~(\ref{eq:J_k_of_delta_k}) involves high-dimensional microscopic dynamical variables $\vec{J}_k^{\tsk}$. It can be projected onto the low-dimensional space of order parameters by multiplying both sides of Eq.~(\ref{eq:J_k_of_delta_k}) with $\vec{B}_n^{\tsk}$ and $\vec{J}_l^{\tsk}$, leading to the evolution of two sets of order parameters,
\begin{align}
    R_{kn}^{t} - R_{kn}^{t-1} & = \frac{\eta_{J}}{NV}\sum_{\nu=1}^V \frac{h_{k}^{\n} y_{n}^{\n}}{\sqrt{K}}, \label{eq:R_diff} \\
    Q_{kl}^{t}-Q_{kl}^{t-1} & = \frac{\eta_{J}}{NV} \sum_{\nu=1}^V \bigg( \frac{h^{\n}_{k}x_{l}^{\n}}{\sqrt{K}} + \frac{h^{\n}_{l}x_{k}^{\n}}{\sqrt{K}} \bigg) \nonumber \\
    & \quad + \frac{\eta_{J}^{2}}{NV^{2}}\sum_{\nu=1}^V\frac{h^{\n}_{k} h^{\n}_{l}}{K}, \label{eq:Q_diff}
\end{align}
where we have made use of the fact that $(\vec{\xi}^{\tsk, \nu})^{\top} \cdot \vec{\xi}^{\tsk, \nu'} = N \delta_{\nu, \nu'}$ almost surely in the limit $N \to \infty$.\footnote{Here, $\delta_{\nu, \nu'}$ stands for the Kronecker delta function.}

Traditional statistical mechanics analysis of online learning arrives at a similar set of dynamical equations of order parameters, and proceeds to examine their deterministic evolution assuming the self-averaging property of the order parameters. 
For our meta-learning problem, it is a bit more tricky to do that, as the right-hand sides of the Eqs.~(\ref{eq:R_diff}) and (~\ref{eq:Q_diff}) have three sources of randomness, (i) from the local fields $( \vec{x}^{\m}, \vec{y}^{\m} )$ through the stochastic inputs $\{ \vec{\xi}^{\m}\}$, (ii) from the local fields $( \vec{x}^{\n}, \vec{y}^{\n} )$ through the stochastic inputs $\{ \vec{\xi}^{\n}\}$, and (iii) from the choice of the $\tsk$th teacher's task vector $\vec{u}^{\tsk}$. 
In Appendix~\ref{sec:self_averaging}, we provide a heuristic argument to show that when $P$ and $V$ are large, the dynamics of the order parameters $\{ R_{kn}, Q_{kl} \}$ admit the self-averaging property, i.e., a realization of their random trajectory stays close to their average values. 
Nevertheless, the values of $P$ and $V$ can still be significantly smaller than those required for successful learning in single-task settings.

In the thermodynamic limit $N\rightarrow \infty$, the normalized task index $\alpha = t/N$ can be interpreted as a continuous time variable, leading to the average dynamics of the order parameters
\begin{align}
    \frac{\dd R_{kn}}{\dd \alpha} & = \frac{\eta_{J}}{V\sqrt{K}} \sum_{\nu = 1}^{V} \big\langle h_{k}^{\nu} y_{n}^{\nu} \big\rangle, \label{eq:dR_da} \\
    \frac{\dd Q_{kl}}{\dd \alpha} & = \frac{\eta_{J}}{V\sqrt{K}} \sum_{\nu = 1}^{V} \big[ \big\langle h_{k}^{\nu} x_{l}^{\nu} \big\rangle +  \big\langle h_{l}^{\nu} x_{k}^{\nu} \big\rangle \big] + \frac{\eta_{J}^{2}}{V^{2}K} \big\langle h_{k}^{\nu} h_{l}^{\nu} \big\rangle,\label{eq:dQ_da}
\end{align}
where $\langle \cdots \rangle$ denotes the average over all sources of randomness. 

\subsection{Dynamics of order parameters}
What remains to be done is the disorder averages of various quantities,
where more details are provided in Appendix.~\ref{sec:app:disorder_average}.
By choosing the error activation function $g(x) = \operatorname{erf}(x/\sqrt{2})$ and applying a large $P$ approximation, we can derive closed-form dynamical equations for the order parameters. Notably, this process requires evaluating a few low-dimensional integrals with a Gaussian measure over the local fields $(\vec{x}, \vec{y}) \sim \mathcal{N}(0, C)$,
\begin{align}
    f_{km} = & \big\langle g(x_{k}) g(y_{m}) \big\rangle_{ (\vec{x}, \vec{y})},\label{eq:f_km_def_main} \\
    I_3(k, n, m) & = \big\langle g^{\prime}(x_k) y_n g(y_m) \big\rangle_{ (\vec{x}, \vec{y})  }, \\
    I_3(k, n, i) & = \big\langle g^{\prime}(x_k) y_n g(x_i) \big\rangle_{ (\vec{x}, \vec{y})  }, \\
    I_4(k, l, n, m) & = \big\langle g^{\prime}(x_k) g^{\prime}(x_l) g(y_n) g(y_m) \big\rangle_{ (\vec{x}, \vec{y}) }, \\
    I_4(k, l, n, i) & = \big\langle g^{\prime}(x_k) g^{\prime}(x_l) g(y_n) g(x_i) \big\rangle_{ (\vec{x}, \vec{y}) }, \\
    I_4(k, l, i, j) & = \big\langle g^{\prime}(x_k) g^{\prime}(x_l) g(x_i) g(x_j) \big\rangle_{ (\vec{x}, \vec{y}) },
\end{align}
where we adhere to the convention that the indices $i,j,k,l$ enumerate the neurons in the hidden layers of the student network, and the indices $n,m$ enumerate the neurons in the hidden layers of the teacher network. All the integrals $f_{km}, I_3, I_4$ are functions of the order parameters $\{Q_{kl}, R_{kn}, T_{mn} \}$. Their detailed expressions can be found in Ref.~\cite{SaadPRE1995}.

Collecting the above results and leveraging the task vector distribution $\vec{u}^{\tsk} \sim \mathcal{N}(0, I_M)$, we derive that the equation of motion of the averaged order parameters $R_{kn}$ can be expressed as
\begin{align}
    \frac{\dd R_{kn}}{\dd \alpha} 
    & = \frac{\eta_{J}\eta_{w}}{KM}\sum_{m=1}^{M}f_{km}I_{3}(k, n, m) \nonumber \\
    & \quad -  \frac{\eta_{J}\eta_{w}^{2}}{K^{2}M}\sum_{i=1}^{K}\sum_{m=1}^{M}f_{in}f_{km}I_{3}(k, n, i), \label{eq:dR_da_avg}
\end{align}
and the equation of motion of the averaged order parameters $Q_{kl}$ can be expressed as
\begin{widetext}
\begin{align}
    \frac{\dd Q_{kl}}{\dd \alpha} & = \frac{\eta_{J}\eta_{w}}{KM}\sum_{n=1}^{M}f_{kn}I_{3}(k, l, n) - \frac{\eta_{J}\eta_{w}^{2}}{K^{2}M}\sum_{i=1}^{K}\sum_{n=1}^{M}f_{in}f_{kn}I_{3}(k, l, i) \nonumber \\
    & \quad + \frac{\eta_{J}\eta_{w}}{KM}\sum_{n=1}^{M}f_{kn}I_{3}(l, k, n) -  \frac{\eta_{J}\eta_{w}^{2}}{K^{2}M}\sum_{i=1}^{K}\sum_{n=1}^{M}f_{in}f_{kn}I_{3}(l, k, i) \nonumber \\
    & \quad + \frac{\eta_{J}^{2}\eta_{w}^{2}}{VK^{2}M^{2}}\sum_{m,n=1}^{M} \bigg\{ f_{km}f_{lm}I_{4}(k, l, n, n) + \big[ f_{kn}f_{lm}  + f_{km}f_{ln} \big] I_{4}(k, l, n, m)  \bigg\}  \nonumber \\
    & \quad - \frac{2\eta_{J}^{2}\eta_{w}^{3}}{VK^{3}M^{2}}\sum_{i=1}^{K}\sum_{m,n=1}^{M} \bigg\{ \big[f_{in}f_{km}f_{lm} + f_{im}f_{kn}f_{lm} + f_{im}f_{km}f_{ln}\big]I_{4}(k, l, n, i) \bigg\} \nonumber \\
    & \quad + \frac{\eta_{J}^{2}\eta_{w}^{4}}{VK^{4}M^{2}}\sum_{i, j=1}^{K}\sum_{m,n=1}^{M}\bigg\{ \big[ f_{in}f_{jn}f_{km}f_{lm} + f_{in}f_{jm}f_{kn}f_{lm} + f_{in}f_{jm}f_{km}f_{ln} \big]I_{4}(k, l, i, j) \bigg\}. \label{eq:dQ_da_avg} 
\end{align}
\end{widetext}

Finally, the meta-generalization error can be computed as
\begin{align}
    \epsilon_g^{\text{meta}} & = \frac{1}{2M}\sum_{n=1}^{M}f_{nn}  - \frac{\eta_w}{KM}\sum_{k =1}^{K}\sum_{n=1}^{M}f_{kn}^{2} \nonumber \\ 
    & \quad + \frac{\eta_w^2}{2K^2 M}\sum_{k,l=1}^{K}\sum_{n=1}^{M}f_{kn}f_{ln}f_{kl}. \label{eq:e_g_meta_avg}
\end{align}

Equations.~(\ref{eq:dR_da_avg}), (\ref{eq:dQ_da_avg}) and (\ref{eq:e_g_meta_avg}) are the major theoretical results of this work. Solving the ordinary differential equations~(\ref{eq:dR_da_avg}) and (\ref{eq:dQ_da_avg}) with initial conditions for $\{ R_{kn}, Q_{kl} \}$ at $\alpha = 0$, we obtain a macroscopic description of the learning dynamics through the order parameters.

\section{Results}\label{sec:results}

\subsection{Comparison to simulation}\label{sec:compare}
Before analyzing various learning scenarios, we compare the theory of meta-learning dynamics developed in Sec.~\ref{sec:teacher_student} to numerical experiments as a sanity check of the theoretical derivation.
For this purpose, we adopt the same task and data generation processes as in Sec.~\ref{sec:teacher} to conduct numerical simulations. The representation mapping of the meta-teacher network is generated as $\vec{B}_{n} \sim \mathcal{N}(0, \frac{1}{N} I_N)$. 
At time $\alpha = 0$, the representation mapping of the meta-learner is also initialized as $\vec{J}_{k} \sim \mathcal{N}(0, \frac{1}{N} I_N)$. Note that $\vec{J}_k(\alpha = 0)$ and $\vec{B}_n$ are statistically independent. The meta-learner is then trained using the FO-ANIL algorithm on the tasks generated by the meta-teacher. In this way, the quantities $R_{kn}$ and $Q_{kl}$ at the initial time corresponding to the simulation setup can be computed; they will be set as the initial conditions of the ordinary differential equations~(\ref{eq:dR_da_avg}) and (\ref{eq:dQ_da_avg}), which remain to be solved to obtain the theoretical predictions.

\begin{figure}[!htbp]
    \centering
    \includegraphics[scale=0.85]{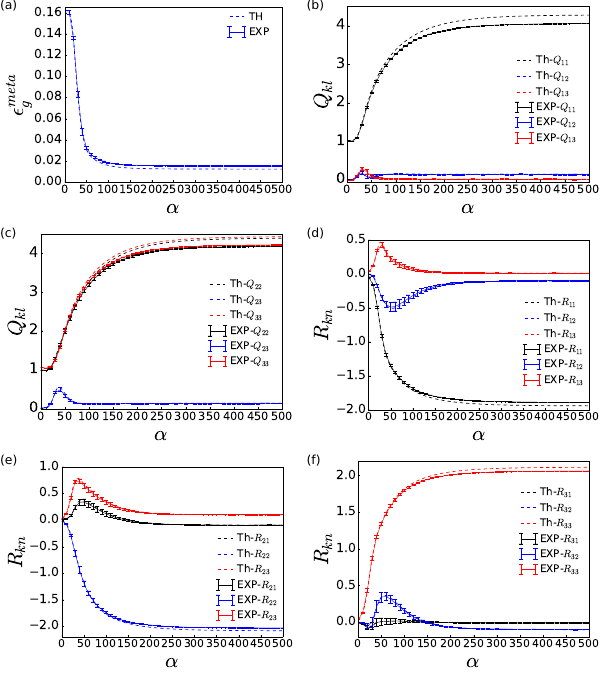} 
    \caption{Comparison between theoretical predictions (marked by ``$\text{TH}$'') and simulated experiments (marked by ``$\text{Exp}$'') of the online meta-learning dynamics under consideration. The simulation setup is outlined in Sec.~\ref{sec:compare}, with system parameters $N=1000, K=M=3, P=V=100, \eta_J=6, \eta_w=4$.     
    The simulations are averaged over $10$ independent runs, each with different realizations of training tasks and datasets, while maintaining the same initial conditions. The error bars represent one standard deviation across these $10$ trials.
    (a) The meta-generalization error $\epsilon_g^{\text{meta}}$ as a function of the normalized task index $\alpha = \tsk / N$. Since directly evaluating $\epsilon_g^{\text{meta}}$ by testing on a large dataset is time-consuming, we instead prepare a small test dataset at each time step and compute a moving average of $\epsilon_g^{\text{meta}}$ over time to smooth out fluctuations for each trial, with a sliding window $\Delta \alpha = 0.05$. 
    Panels (b) and (c) depict the dynamic behavior of $Q$, where solid lines indicate experimental observations and dashed lines show theoretical results.
    Panels (d), (e), and (f) illustrate the dynamical evolution of $R$, with solid lines representing experimental data and dashed lines corresponding to theoretical predictions.}
    \label{fig:E_D}
\end{figure}

The other system parameters are set as $N = 1000, K = M = 3, P = V = 100, \eta_J = 6, \eta_w = 4$. The results for numerical simulations and the corresponding theoretical predictions are shown in Fig.~\ref{fig:E_D}.
It can be observed that the theoretical curves of $\epsilon_g^{\text{meta}}$ and $\{ R_{kn} \}$ closely match the experimental data, which suggests that the theoretical model effectively captures the underlying phenomena observed in the experiments. The meta-generalization error $\epsilon_g^{\text{meta}}$ decrease as the training process evolves, suggesting that the meta-learner effectively learns the knowledge of the meta-teacher.
On the other hand, the deviation of theory and simulation for $\{ Q_{kl} \}$ is a bit more prominent for the chosen set of system parameters. In Appendix~\ref{sec:self_averaging}, we provide an argument and further numerical results to support that the order parameters $\{ Q_{kl} \}$ are subject to larger statistical error when $V$ is finite and $\eta_J$ and $\eta_w$ are large.
When $P, V$ are sufficiently large, we expect that the theory provides an accurate description of the dynamics of both sets of order parameters $\{ R_{kn} \}$ and $\{ Q_{kl} \}$. 

\subsection{Meta-representation learning}\label{sec:representation_KM3_case}

In this section, we elaborate on the mechanism driving the meta-generalization ability of the meta-learner which has already been observed in Fig.~\ref{fig:E_D}, by focusing on a case study with $K = M = 3$.

A sufficient condition for successful meta-learning is that the meta-learner's representation mapping $\{ \vec{J}_k \}$ aligns with the meta-teacher's representation mapping $\{ \vec{B}_n \}$, up to the permutation and reflection symmetries of the hidden units.\footnote{By reflection symmetry, we mean that the solution of $\vec{J}_k$ is allow to aligned with the opposite direction of $\vec{B}_n$ for some hidden teacher neuron $n$ (suppose that $\vec{J}_k = - \vec{B}_n$); this is because the chosen activation function $g(x) = \operatorname{erf}(x / \sqrt{2})$ is an odd function, so that setting $w^{\tsk}_k = - u^{\tsk}_n$ enables the $\tsk$th learner to learn the output component contributed by the $n$th hidden unit of the $\tsk$th teacher.}
This can be measured by monitoring the overlaps $\{ R_{kn} := \vec{J}_k^{\top} \cdot \vec{B}_n \}$ between the meta-learners' and meta-teachers' hidden units, and their cosine similarity defined by
\begin{align}
    \rho_{kn} = \frac{ \vec{J}_k^{\top} \cdot \vec{B}_n }{ \sqrt{\vec{J}_k^{\top} \cdot \vec{J}_k} \sqrt{\vec{B}_n^{\top} \cdot \vec{B}_n} } = \frac{R_{kn}}{\sqrt{Q_{kk}T_{nn}}}.
\end{align}
The cosine similarity, $\rho_{kn}$, lies within the interval $[-1, 1]$, where $|\rho_{kn}| = 1$ indicates a perfect parallel alignment between $\vec{J}_k$ and $\vec{B}_n$.

Suppose that a successful alignment occurs with $R_{kn} \propto \delta_{k, n}$ (i.e., $\vec{J}_k \propto \vec{B}_n$); then, based on Eqs.~(\ref{eq:w_by_u_f}) and (\ref{eq:f_km_by_arcsin}), the $k$th hidden-to-output weight of the $\tsk$th learner will, on average, be determined by the $n$th hidden-to-output weight of the $\tsk$th teacher (i.e., $\langle w^{\tsk}_k \rangle \propto u^{\tsk}_{n}$). If the learning rate $\eta_w$ is properly set and the order parameters $\{ Q_{kk} \}$ are accurately learned, the $\tsk$th learner could potentially reproduce the exact output of the $\tsk$th teacher. Consequently, we will focus on the similarity between the representation mappings $\mat{J}$ and $\mat{B}$ and examine their relationship with meta-generalization performance.

\begin{figure}[!htbp]
    \centering
    \includegraphics[scale=0.8]{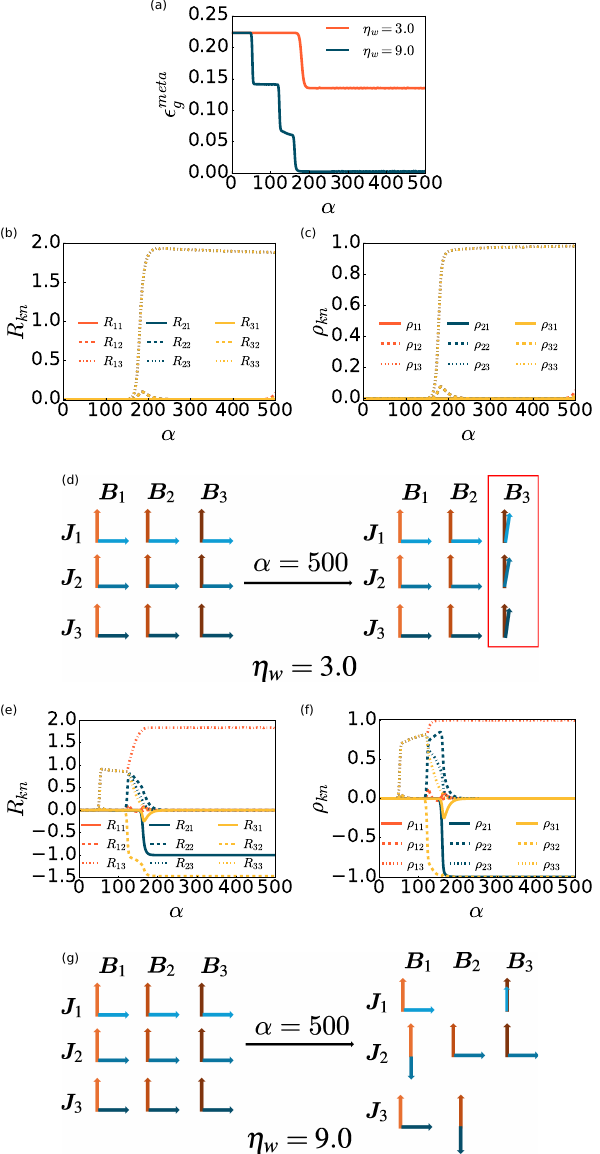} 
    \caption{A case study of meta-representation learning. The system parameters are $ K = M = 3, \eta_J = 3, T_{mn} = m \delta_{m,n} $, and the initial conditions are $ Q_{kl} = \frac{1}{2}\delta_{k,l}, R_{kn} = 10^{-12} $. 
    (a) $\epsilon_g^{\text{meta}}$ vs $\alpha$. 
    (b) $R_{kn}$ vs $\alpha$ for $\eta_w = 3$. 
    (c) $\rho_{kn}$ vs $\alpha$ for $\eta_w = 3$. 
    (d) Pictorial illustration of the outcome of the meta-representation learning for $\eta_w = 3$.
    (e) $R_{kn}$ vs $\alpha$ for $\eta_w = 9$. 
    (f) $\rho_{kn}$ vs $\alpha$ for $\eta_w = 9$. 
    (g) Pictorial illustration of the outcome of the meta-representation learning for $\eta_w = 9$.}
    \label{fig:represatation}
\end{figure}

Inspired by \cite{SaadPRE1995}, we set the system parameters as $K=M=3, \eta_J=3, T_{mn} = m \delta_{m, n}$ and consider the initial conditions $Q_{kl} = \frac{1}{2}\delta_{kl}, R_{kn} = 10^{-12}$. The results for two different choices of $\eta_w$ are shown in Fig.~\ref{fig:represatation}. From Fig.~\ref{fig:represatation}(a), we observe that $\epsilon_g^{\text{meta}}$ decreases with increasing $\alpha$ for both cases, indicating the improvement of the meta-generalization ability of the meta-learner.

For $\eta_w = 3$, $\epsilon_g^{\text{meta}}$ experiences an initial drop and remains on a plateau for the duration of the analysis, up to $\alpha = 500$.
The initial drop occurs at around $\alpha = 180$, at which time $R_{13}$, $R_{23}$, and $R_{33}$ (or $\rho_{13}$, $\rho_{23}$, and $\rho_{33}$) start to grow from zero to a similar value as observed in Fig.~\ref{fig:represatation}(b) and (c).
This implies that all the meta-learner's hidden units align with only one hidden unit ($n=3$) of the meta-teacher network [as illustrated in Fig.~\ref{fig:represatation}(d)], which is a suboptimal solution.

On the other hand, for $\eta_w = 9$, $\epsilon_g^{\text{meta}}$ undergoes three distinct drops, ultimately decreasing to a value close to zero. Analyzing the behaviors of $R_{kn}$ and Fig.~\ref{fig:represatation}(e), we observe that $R_{13}$, $R_{32}$, and $R_{21}$ progressively develop and eventually converge to distinct nonzero values, while the remaining order parameters $\{ R_{kl} \}$ approach zero. A similar pattern is seen for the cosine similarities $\{ \rho_{kl} \}$ shown in Fig.~\ref{fig:represatation}(f).
At around $\alpha = 40$, the meta-learner gets into the symmetric suboptimal state where all its hidden units learn the single hidden unit ($n=3$) of the meta-teacher, indicated by the growth of $R_{13}, R_{23}$ and $R_{33}$. But eventually, the meta-learner is able to escape this symmetric state by further differentiating its hidden units.
In the final stage, the hidden units in the meta-learner specialized to different hidden units in the meta-teacher network, i.e., $\vec{J}_1 \propto \vec{B}_3, \vec{J}_2 \propto -\vec{B}_1, \vec{J}_3 \propto -\vec{B}_2$ as illustrated in Fig.~\ref{fig:represatation}(g). Up to the permutation and reflection symmetries, the meta-learner's representation mapping perfectly align with that of the meta-teacher.
Consequently, the meta-generalization error can ultimately be reduced to a value close to zero.

The symmetry breaking and specialization transition of hidden units are hallmarks of the traditional teacher-student model for single-task learning, which drives the generalization ability of the learning machines~\cite{SaadPRE1995}. Here, we show that the same phenomena can also occur in the meta-learning setting, and that they can be achieved under the learning dynamics dictated by the FO-ANIL algorithm. Moreover, it is shown that the success of meta-representation learning does correspond to good meta-generalization ability.

Although the phenomenology here is similar to that of single-task learning, the fact that the simple FO-ANIL meta-learning algorithm successfully learns the underlying representation is remarkable, since the substantial variability in hidden-to-output weights across different task-specific teachers introduces significant noise, making the learning of this representation considerably more challenging.

\subsection{The role of learning Rates}\label{sec:learning_rate}
As seen in Sec.~\ref{sec:representation_KM3_case}, it is crucial to determine the appropriate learning rate $\eta_w$ for the effectiveness of meta-learning. We expect that the choice of $\eta_J$ also matters.
Here we study the effects of learning rates in greater detail within our theoretical framework. 

\begin{figure} 
    \centering
    \includegraphics[scale=1.0]{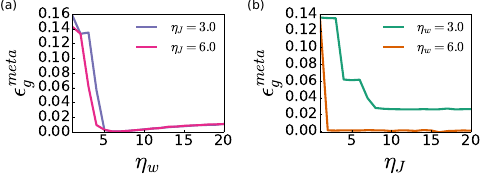} 
    \caption{Effect of learning rates for meta-generalization. The system parameters and initial conditions are the same as those in Sec.~\ref{sec:representation_KM3_case}. (a) $\epsilon_g^{\text{meta}}$ vs $\eta_w$ when fixing $\eta_J$. (b) $\epsilon_g^{\text{meta}}$ vs $\eta_J$ when fixing $\eta_w$. }
    \label{fig:learning_rate}
\end{figure}

We set the same system parameters and initial conditions as those in the case study in Sec.~\ref{sec:representation_KM3_case}, except that the learning rates $\eta_J$ and $\eta_w$ can vary. The meta-learning dynamics is evolved up to time $\alpha = 500$, at which the final value of $\epsilon_g^{\text{meta}}$ is extracted for each choice of learning rates. The results are shown in Fig.~\ref{fig:learning_rate}.
It can be observed from Fig.~\ref{fig:learning_rate}(a) that for a particular learning rate $\eta_{J}$, the final meta-generalization error first decreases rapidly and then increases gently when $\eta_{w}$ increases. 
There is an optimal value of $\eta_w$, but increasing it to a relatively large value does not deteriorate the performance much.
In Fig.~\ref{fig:learning_rate}(b), we observe that for a particular learning rate $\eta_{w}$, the final meta-generalization error decreases and saturates when $\eta_{J}$ increases. The case with $\eta_{w} = 6$ yields a better generalization performance than the case with $\eta_{w} = 3$ for various choices of $\eta_J$. 

\subsection{Overparameterization}
In earlier sections, we examined learning scenarios where $K = M$. In practice, however, the meta-learner typically does not have access to the meta-teacher's architecture and may set $K$ differently from $M$. It is well recognized in deep learning practice that overparameterized neural networks (those with more parameters than necessary) do not necessarily lead to overfitting~\cite{Zhang2017}. On the contrary, when properly trained, overparameterization can enhance system performance~\cite{Nakkiran2020}.
Previous studies have suggested that overparameterized models, particularly under gradient-based algorithms, are beneficial for optimization and can provide implicit regularization, leading to improved generalization~\cite{Nguyen2017, Zhu2019, Neyshabur2018, Arora2019}. Despite these insights, a comprehensive understanding of overparameterization remains elusive, indicating the need for case-by-case analysis.

\begin{figure} 
    \centering
    \includegraphics[scale=1.0]{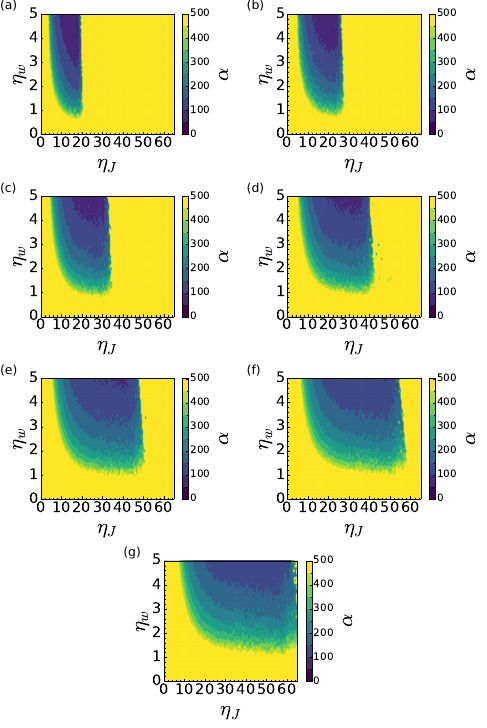} 
    \caption{The values of $\tilde{\alpha}$ needed to reach $\epsilon_g^{\text{meta}} = 0.01$ under different choices of the learning rates $\eta_w$ and $\eta_J$. The yellow region indicates that $\epsilon_g^{\text{meta}}$ fails to drop to the level $0.01$ within the time window $\alpha_{\text{final}} = 450$ considered.
    The system parameters and initial conditions are set as $M = 3, T_{mn} = m \delta_{m,n}, Q_{kl} = \frac{1}{2} \delta_{k,l}, R_{k,n} = 10^{-12}$.
    Panels (a) to (g) correspond to $K = 3, 4, 5, 6, 7, 8, 9$, respectively.}
    \label{fig:Over}
\end{figure}

In our teacher-student framework, overparameterization can be introduced by setting $K > M$.
For a case study, we fix the number of hidden units in the meta-teacher network at $M = 3$ and increase the number of hidden units $K$ in the meta-learner. The other system parameters and initial conditions are set as $T_{mn} = m \delta_{m,n}, Q_{kl} = \frac{1}{2} \delta_{k,l}, R_{k,n} = 10^{-12}$. The meta-learning dynamics is evolved up to time $\alpha_{\text{final}} = 450$, and during the meta-training process, we record the time $\tilde{\alpha}$ at which $\epsilon_g^{\text{meta}}$ decreases to $0.01$.
The results for $\tilde{\alpha}$ under different choices of the learning rates $\eta_w$ and $\eta_J$ are shown in Fig.~\ref{fig:Over}, with panels from (a) to (g) corresponding to $K = 3, 4, 5, 6, 7, 8, 9$, respectively. 
As the color shifts from blue to yellow, the number of tasks required to reach $\epsilon_g^{\text{meta}} = 0.01$ gradually increases, with the yellow region indicating that $\epsilon_g^{\text{meta}}$ fails to drop to the level $0.01$ within the time window $\alpha_{\text{final}} = 450$ considered. 

As $K$ increases, both the minimal values of $\eta_w$ and $\eta_J$ required for the meta-learner to achieve $\epsilon_g^{\text{meta}} = 0.01$ before $\alpha_{\text{final}} = 450$ also increase. This can be understood by referring to Eqs.~\ref{eq:dR_da}, \ref{eq:dQ_da} and \ref{eq:w_by_u_f}, which show that both $\eta_w$ and $\eta_J$ are scaled by the factor $\frac{1}{\sqrt{K}}$. Therefore, larger values of $\eta_w$ and $\eta_J$ are necessary to maintain a comparable pace of the dynamics when $K$ increases.
More interestingly, as $K$ increases, the nonyellow region expands, indicating that in more overparameterized models, less stringent choices of the learning rates (particularly $\eta_J$) can be used to achieve a strong level of meta-generalization for a given number of tasks. This echos previous findings that overparameterization facilitates optimization.

\begin{figure}[!htbp]
    \centering
    \includegraphics[scale=1.0]{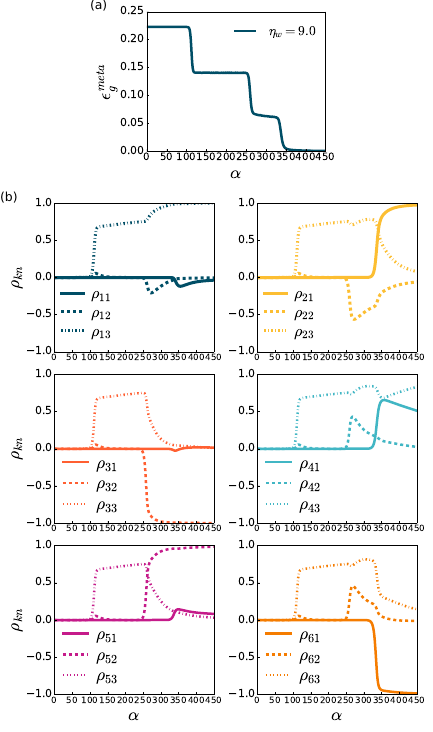} 
    \caption{Meta-learning dynamics of an overparameterized model with $M=3, K=6$. The other system parameters and initial conditions are $\eta_J=3, \eta_w=9, T_{mn} = m \delta_{m,n}, Q_{kl} = \frac{1}{2} \delta_{k,l}, R_{kn} = 10^{-12}$. 
    (a) $\epsilon_g^{\text{meta}}$ vs $\alpha$.
    (b) $\rho_{kn}$ v.s. $\alpha$. 
    It can be observed that $|\rho_{21}|, |\rho_{61}|$ converge to one, indicating that the second and sixth hidden units in the meta-learner have specialized to the first hidden unit in the meta-teacher network. Similarly, $|\rho_{32}|, |\rho_{52}|$ converge to one, which suggests that the third and fifth hidden units in the meta-learner have specialized to the second hidden unit in the meta-teacher. Finally, $|\rho_{13}|, |\rho_{43}|$ converge to one, which indicates that the first and fourth hidden units in the meta-learner have specialized to the third hidden unit in the meta-teacher.}
    \label{fig:Over_repre}
\end{figure}

To investigate the representation learning mechanism in overparameterized models, we plotted the evolution of $\epsilon_g^{\text{meta}}$ and $\rho$ for a case with $K=6, \eta_J = 3, \eta_w = 9$ in Fig.~\ref{fig:Over_repre}. 
The results indicate that each hidden unit in the meta-teacher network has been effectively learned by two hidden units in the meta-learner, a phenomenon which has been observed in the single-task scenario when training both layers~\cite{Goldt2019}. This added flexibility in the overparameterized meta-learner may provide an optimization advantage, allowing it to better capture the meta-teacher.

\subsection{Variability of teacher representation mapping} \label{sec:dB}
In previous sections, we have primarily focused on the scenario that different task-specific teachers share the same representation mapping $\mat{B}$, derived from the meta-teacher network.
Here, we relax this assumption by allowing task-specific teachers to have different representation mappings, while maintaining a certain level of similarity.
To model this scenario, we define the input-to-hidden weights of the $\tsk$th teacher network as
\begin{align}
    \mat{B}^{\tsk} = \gamma\mat{B} + \sqrt{1-\gamma^{2}}\Delta\mat{B}^{\tsk},
\end{align}
where $\mat{B}$ is the common input-to-hidden weights of the meta-teacher shared across all tasks, while $\Delta \mat{B}^{\tsk}$ is specific to the $\tsk$th task. The parameter $\gamma \in [0, 1]$ controls the variability of the teacher weights across tasks. We assume that $\Delta\mat{B}^{\tsk}$ is drawn from a certain distribution $p(\Delta \mat{B})$.
When $\gamma$ is close to one, the contribution of the component $\sqrt{1-\gamma^{2}}\Delta\mat{B}^{\tsk}$ is small, which allows us to perform a series expansion of the activation $ g( \vec{B}_m^{\tsk \top} \cdot \vec{\xi} ) = g( \gamma \vec{B}_m^{\top} \cdot \vec{\xi} + \sqrt{1 - \gamma^2} \Delta\vec{B}_m^{\tsk \top} \cdot \vec{\xi} )$ around the value $\gamma \vec{B}_m^{\top} \cdot \vec{\xi}$.
In this manner, we can derive the dynamical equations for the order parameters using the large $\gamma$ approximation. Further details of the derivation are given in Appendix~\ref{sec:details_dB}. 

Figure~\ref{fig:dB}(a) illustrates the evolution of $\epsilon_g^{\text{meta}}$ during the meta-training process for various values of $\gamma$. It demonstrates that even when $\gamma < 1$, indicating variability in the teacher's representation mapping, the FO-ANIL algorithm is still capable of effectively reducing the meta-generalization error, though not as optimally as in the case with $\gamma = 1$.
Figure~\ref{fig:dB}(b) depicts the effect of $\eta_w$ on $\epsilon_g^{\text{meta}}$ after a certain time window, which reassures the efficacy of the meta-learning algorithm for different choices of the learning rates.
Similar to the findings in Sec.~\ref{sec:learning_rate}, there is an optimal value of $\eta_w$ for each case, but increasing it to a relatively large value does not significantly impair performance.

\begin{figure}[!htbp]
    \centering
    \includegraphics[scale=1]{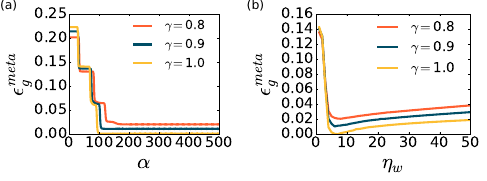}
    \caption{Meta-generalization in the presence of variability of teacher representation mapping. The system parameters and initial conditions are $K = M = 3, \eta_J = 6, T_{mn} = m\delta_{m,n}, Q_{kl} = \frac{1}{2}\delta_{k,l}, R_{kn} = 10^{-12}$. 
    (a) $\epsilon_g^{\text{meta}}$ as a function of $\alpha$ for $\eta_w = 8$.
    (b) $\epsilon_g^{\text{meta}}$ at the final time $\alpha = 500$ as a function of $\eta_w$ for $\eta_J = 6$.}
    \label{fig:dB}
\end{figure}

\section{Conclusion}\label{sec:conclusion}

In summary, we conducted a detailed theoretical analysis to gain insights into the dynamics of meta-learning on two-layer networks under gradient-based algorithms, with a particular focus on the first-order almost no inner loop (FO-ANIL) algorithm. 

By developing a teacher-student model tailored to our learning scenarios and utilizing tools from statistical physics, we were able to simplify the complex, high-dimensional learning dynamics into an effective low-dimensional description through macroscopic order parameters. This approach allowed us to identify the key factors that influence the learning process. 
We discovered that meta-generalization is achieved through effective meta-representation learning by the meta-learner, driven by specialization transitions within its hidden units. Before these transitions occur, the system may remain in a suboptimal state with high meta-generalization error for an extended period, where multiple hidden units of the meta-learner focus on learning from a single hidden unit of the meta-teacher, while neglecting others. As meta-training continues, the meta-learner gradually gathers sufficient signals to escape this symmetric phase, leading to the specialization of its hidden units to those of the meta-teacher. 
These phenomena are similar to the teacher-student scenario in a single-task setting, but it is far more remarkable that the teacher's representation mapping can be successfully learned under the simple FO-ANIL algorithm. The large variability in the hidden-to-output weights in different task-specific teachers introduces significant noise, making the learning of representation mappings much more challenging.

For a fixed number of tasks (i.e., within a specific time window of the dynamics), we found that choosing appropriate learning rates ($\eta_w$ for inner optimization and $\eta_J$ for outer optimization) is essential for improving the model's meta-generalization ability; neither rate should be too small. 
Additionally, we explored overparameterized meta-learners which have more hidden units than the meta-teachers, and showed the flexibility they offer in selecting learning rates. 
Future work could investigate theoretically optimal learning rates for the meta-learner using the methods proposed in~\cite{Saad1997}.
We also demonstrated that meta-learning algorithms can handle situations where the input-to-hidden weights of each task-specific teacher deviate from those of the meta-teacher.
The proposed framework can also accommodate other learning scenarios, such as the effect of explicit regularization, other activation functions, and so on, which we briefly discuss in Appendix~\ref{sec:L2reg} and \ref{sec:linear_act_func}. 
We believe that this work offers a solid framework and a wealth of examples for the statistical-mechanical analysis of meta-learning, with potential for extension to explore other interesting meta-learning scenarios.

\begin{acknowledgments}
We thank David Saad for fruitful discussions.
H.W. and B.L. acknowledge support from the National Natural Science Foundation of China (Grant No. 12205066), the Guangdong Basic and Applied Basic Research Foundation (Grant No. 2024A1515011775), the Shenzhen Start-Up Research Funds (Grant No. BL20230925) and the start-up funding from the Harbin Institute of Technology, Shenzhen (Grant No. 20210134). C.T.Y. acknowledges support from the Guangdong Basic and Applied Basic Research Foundation (Grant No. 2214050004792), the National Natural Science Foundation of China (Grant No. 1217040938), and the Shenzhen Municipal Science and Technology projects (Grant No. 202001093000117).
\end{acknowledgments}

\section*{References}
\bibliography{reference}

\clearpage
\begin{widetext}
\appendix

\section{More Details on Disorder Averages} \label{sec:app:disorder_average}
Here we provide more details on the disorder averages of the order parameters.
As an illustrating example, we consider the computation of $\langle h_{k}^{\nu} y_{n}^{\nu} \rangle$ in Eq.~(\ref{eq:dR_da}), by invoking the three types of disordered averages
\begin{align}
    & \quad \;\; \big\langle h_{k}^{\nu} y_{n}^{\nu} \big\rangle \label{eq:hk_yn} \\
    & = \bigg\langle \bigg[ \sum_{m=1}^{M}\frac{u_{m} g(y_{m}^{\nu})}{\sqrt{M}}-\sum_{i=1}^{K}\frac{w_{i} g(x_{i}^{\nu})}{\sqrt{K}} \bigg]  w_{k} g^{\prime}(x_{k}^{\nu}) y_{n}^{\nu} \bigg\rangle \nonumber \\
    & = \frac{1}{\sqrt{M}} \sum_{m=1}^{M} \bigg\langle u_m \big\langle w_k \big\rangle_{ \{ \vec{\xi}^{\mu} \} } \big\langle g^{\prime}(x_k^{\nu}) y_n^{\nu} g(y_m^{\nu}) \big\rangle_{ \{ \vec{\xi}^{\nu} \} } \bigg\rangle_{ \vec{u} } \nonumber \\
    & \quad \;\; - \frac{1}{\sqrt{K}} \sum_{i=1}^{K} \bigg\langle \big\langle w_i w_k \big\rangle_{ \{ \vec{\xi}^{\mu} \} } \big\langle  g^{\prime}(x_k^{\nu}) y_n^{\nu} g(x_i^{\nu}) \big\rangle_{ \{ \vec{\xi}^{\nu} \} } \bigg\rangle_{ \vec{u} }. \nonumber
\end{align}

The average $\langle w_k \rangle_{ \{ \vec{\xi}^{\mu} \} }$ only involves the training inputs at time $\alpha$, which can be computed as 
\begin{align}
    \big\langle w_k \big\rangle_{ \{ \vec{\xi}^{\mu} \} } & = \frac{\eta_{w}}{P \sqrt{MK} }\sum_{\mu=1}^P\sum_{m=1}^{M} u_{m} \big\langle g(x_{k}^{\mu}) g(y_{m}^{\mu}) \big\rangle_{ \{ \vec{\xi}^{\mu} \} } \nonumber \\
    & = \frac{\eta_{w}}{\sqrt{K}} \frac{1}{\sqrt{M}} \sum_{m=1}^{M} u_{m} f_{km}, \label{eq:w_by_u_f} \\
    f_{km} := & \big\langle g(x_{k}) g(y_{m}) \big\rangle_{ (\vec{x}, \vec{y}) \sim \mathcal{N}(0, C) }. \label{eq:f_km_def}
\end{align}
For the error activation function $g(x) = \operatorname{erf}(x/\sqrt{2})$ which we have specified, Eq.~(\ref{eq:f_km_def}) involves only a two-dimensional integration with a Gaussian measure, which admits a close-form expression as
\begin{align}
    f_{km} = \frac{2}{\pi}\arcsin\frac{R_{km}}{\sqrt{1 + Q_{kk}}\sqrt{1+ T_{mm}}}. \label{eq:f_km_by_arcsin}
\end{align}
For the term $\langle w_i w_k \rangle_{ \{ \vec{\xi}^{\mu} \} }$ in Eq.~(\ref{eq:hk_yn}), we employ the following approximation to simplify the calculation
\begin{align}
    \langle w_i w_k \rangle_{ \{ \vec{\xi}^{\mu} \} } & \approx \langle w_i \rangle_{ \{ \vec{\xi}^{\mu} \} } \langle w_k \rangle_{ \{ \vec{\xi}^{\mu} \} } \nonumber \\
    & = \frac{\eta_w^2}{MK} \sum_{m,n=1}^{M} u_n u_m f_{in} f_{km},
\end{align}
where the approximation error is of order $O(\frac{1}{P})$ (see Appendix.~\ref{sec:ww_approx}).

The terms like $\langle g^{\prime}(x_k^{\nu}) y_n^{\nu} g(y_m^{\nu}) \rangle_{ \{ \vec{\xi}^{\nu} \} }$ in Eq.~(\ref{eq:hk_yn}) can also be evaluated analytically as shown in Ref.~\cite{SaadPRE1995}. We write the corresponding result using the short-hand notation introduced in \cite{SaadPRE1995}
\begin{align}
    I_3(k, n, m) & = \big\langle g^{\prime}(x_k^{\nu}) y_n^{\nu} g(y_m^{\nu}) \big\rangle_{ (\vec{x}^{\nu}, \vec{y}^{\nu}) \sim \mathcal{N}(0, C) }, \\
    I_3(k, n, i) & = \big\langle g^{\prime}(x_k^{\nu}) y_n^{\nu} g(x_i^{\nu}) \big\rangle_{ (\vec{x}^{\nu}, \vec{y}^{\nu}) \sim \mathcal{N}(0, C) },
\end{align}
where we adhere to the convention that the indices $i,j,k,l$ enumerate the neurons in the hidden layers of the student network, and the indices $n,m$ enumerate the neurons in the hidden layers of the teacher network. Similar to $f_{mk}$, the outcome of the integral $I_3$ is a function of the order parameters $\{Q_{kl}, R_{kn}, T_{mn} \}$.

For the average over the task vectors $\vec{u}$, we note that they are generated by $\vec{u} \sim \mathcal{N}(0, I_M)$, leading to
\begin{align}
    \langle u_n u_m \rangle_{\vec{u}} = \delta_{n,m}.
\end{align}

As for the evolution of $Q_{kl}$ in Eq.~(\ref{eq:dQ_da}), we need to compute an additional set of averages like $\langle g^{\prime}(x_k^{\nu}) g^{\prime}(x_l^{\nu}) g(y_n^{\nu}) g(y_m^{\nu}) \rangle_{ \{ \vec{\xi}^{\nu} \} }$. It also admits a close-form solution, denoted as
\begin{align}
    & I_4(k, l, n, m) \nonumber \\
    & = \big\langle g^{\prime}(x_k^{\nu}) g^{\prime}(x_l^{\nu}) g(y_n^{\nu}) g(y_m^{\nu}) \big\rangle_{ (\vec{x}^{\nu}, \vec{y}^{\nu}) \sim \mathcal{N}(0, C) }, 
\end{align}
where the detailed expressions can be found in Ref.~\cite{SaadPRE1995}.

\section{Approximation of $\langle w_i w_k \rangle_{ \{ \vec{\xi}^{\mu} \} }$} \label{sec:ww_approx}
When deriving the dynamical equations of the order parameter $R_{kn}$, we have made use of the decorrelation assumption of $\langle w_i w_k \rangle_{ \{ \vec{\xi}^{\mu} \} } \approx \langle w_i \rangle_{ \{ \vec{\xi}^{\mu} \} } \langle w_k \rangle_{ \{ \vec{\xi}^{\mu} \} }$ in the limit of large $P$. Here we briefly outline the justification for this approximation.

We observe that
\begin{align}
    w_{i} w_{k} & = \bigg[\frac{\eta_{w}}{P \sqrt{MK} }\sum_{\mu'=1}^P\sum_{n=1}^{M} u_{n} g(x_{i}^{\mu'}) g(y_{n}^{\mu'}) \bigg] \bigg[\frac{\eta_{w}}{P \sqrt{MK} }\sum_{\mu=1}^P\sum_{m=1}^{M} u_{m} g(x_{k}^{\mu}) g(y_{m}^{\mu}) \bigg], \nonumber \\
    & = \frac{\eta_{w}^{2}}{MK} \sum_{m, n=1}^{M} u_{n} u_{m} \bigg[ \frac{1}{P^2} \sum_{\mu, \mu'=1}^P g(x_{i}^{\mu'}) g(y_{n}^{\mu'}) g(x_{k}^{\mu}) g(y_{m}^{\mu}) \bigg]. \label{eq:wiwk_before_average}
\end{align}

Recall that different example inputs $\{ \vec{\xi}^{\mu} \}$ in the same task are drawn independently from a certain distribution. Therefore, for $\mu' \neq \mu$, $g(x_{i}^{\mu'}) g(y_{n}^{\mu'})$ and $g(x_{k}^{\mu}) g(y_{m}^{\mu})$ can be treated as independent random variables, and their averages can be performed separately
\begin{align}
    \big\langle g(x_{i}^{\mu'}) g(y_{n}^{\mu'}) g(x_{k}^{\mu}) g(y_{m}^{\mu}) \big\rangle_{ \vec{\xi}^{\mu'}, \vec{\xi}^{\mu} } = \big\langle g(x_{i}^{\mu'}) g(y_{n}^{\mu'}) \big\rangle_{ \vec{\xi}^{\mu'} } \times \big\langle g(x_{k}^{\mu}) g(y_{m}^{\mu}) \big\rangle_{ \vec{\xi}^{\mu} }, \quad \forall \mu' \neq \mu.
\end{align}

For $\mu' = \mu$, in principle, we need to deal with a cumbersome four-dimensional integral for computing $\big\langle g(x_{i}^{\mu}) g(y_{n}^{\mu}) g(x_{k}^{\mu}) g(y_{m}^{\mu}) \big\rangle_{ \vec{\xi}^{\mu} }$.

Observe that the summation $\frac{1}{P^2} \sum_{\mu, \mu'=1}^{P} (\cdots)$ in Eq.~(\ref{eq:wiwk_before_average}) has $P^2$ terms, while there are $P(P-1)$ terms satisfying $\mu' \neq \mu$. In the limit where $P$ is large, then the summation $\frac{1}{P^2} \sum_{\mu, \mu'=1}^{P} (\cdots)$ is dominated by the terms of $\mu' \neq \mu$. It allows us to make the approximation
\begin{align}
    \langle w_{i} w_{k} \rangle_{\{ \vec{\xi}^{\mu} \} } 
    & \approx \frac{\eta_{w}^{2}}{MK} \sum_{m, n=1}^{M} u_{n} u_{m} \bigg[ \frac{1}{P^2} \sum_{\mu, \mu'=1}^P \big\langle g(x_{i}^{\mu'}) g(y_{n}^{\mu'}) \big\rangle_{ \vec{\xi}^{\mu'} } \times \big\langle g(x_{k}^{\mu}) g(y_{m}^{\mu}) \big\rangle_{ \vec{\xi}^{\mu} } \bigg] \nonumber \\
    & = \langle w_i \rangle_{ \{ \vec{\xi}^{\mu} \} } \langle w_k \rangle_{ \{ \vec{\xi}^{\mu} \} },
\end{align}
with an approximation error of the order $O(\frac{1}{P})$.

\section{Self-Averaging of the Learning Dynamics in the Limit of Large $P$ and $V$}\label{sec:self_averaging}
In traditional teacher-student analyses of online learning of a single task, the dynamical trajectories of many learning machines exhibit a self-averaging property in the high dimension limit $N \to \infty$~\cite{Reents1998}.
Here, inspired by Ref.~\cite{Engel2001}, we provide a heuristic argument supporting the self-averaging property of the online meta-learning dynamics introduced in the main text, considering the limit of large $P$ and $V$ in addition to the limit of large $N$. Unlike conventional online learning, online meta-learning introduces an additional source of disorder due to the variability in the teachers' task vectors $\{ \vec{u}^{t} \}$.

For illustration, we consider the discrete-time evolution of $R_{kn}$ in Eq.~(\ref{eq:R_diff}). By iterating the dynamics from $t$ to $t + \Delta t$, we obtain
\begin{align}
    R_{kn}^{t + \Delta t} - R_{kn}^{t} & = \frac{\eta_{J}}{\sqrt{K}} \frac{1}{N} \sum_{\tau=1}^{\Delta t} \bigg( \frac{1}{V} \sum_{\nu=1}^V h_{k}^{t+\tau, \nu} y_{n}^{t+\tau, \nu} \bigg), \label{eq:D_R_for_self_average}
\end{align}
where 
\begin{align}
    h^{t+\tau, \nu}_{k} & = \bigg[ \sum_{m=1}^{M}\frac{u_{m}^{t+\tau}g(y_{m}^{t+\tau, \nu})}{\sqrt{M}}-\sum_{i=1}^{K}\frac{w_{i}^{t+\tau}g(x_{i}^{t+\tau, \nu})}{\sqrt{K}} \bigg] w_{k}^{t+\tau}g^{\prime}(x_{k}^{t+\tau, \nu}), \label{eq:h_for_self_average} \\
    w_{k}^{t+\tau} & = \frac{\eta_{w}}{\sqrt{MK}} \sum_{m=1}^{M} u_{m}^{t+\tau} \bigg[ \frac{1}{P} \sum_{\mu=1}^P g(y_{m}^{t+\tau, \mu})g(x_{k}^{t+\tau, \mu}) \bigg], \label{eq:w_for_self_average} \\
    ( \vec{x}^{t+\tau, \mu}, & \vec{y}^{t+\tau, \mu} ) \sim \mathcal{N}(0, C), \\
    ( \vec{x}^{t+\tau, \nu}, & \vec{y}^{t+\tau, \nu} ) \sim \mathcal{N}(0, C), \qquad ( \vec{x}^{t+\tau, \nu}, \vec{y}^{t+\tau, \nu} ) \perp\!\!\!\perp ( \vec{x}^{t+\tau, \mu}, \vec{y}^{t+\tau, \mu} ), \\
    \vec{u}^{t+\tau} & \sim \mathcal{N}(0, I_M).
\end{align}

When $P$ is large, the summation $\frac{1}{P} \sum_{\mu=1}^P g(y_{m}^{t+\tau, \mu})g(x_{k}^{t+\tau, \mu})$ in Eq.~(\ref{eq:w_for_self_average}) converges to its average value $f_{km} = \big\langle g(x_{k}) g(y_{m}) \big\rangle_{ (\vec{x}, \vec{y}) \sim \mathcal{N}(0, C)}$. 

When $V$ is large, the terms $\frac{1}{V} \sum_{\nu=1}^V g(y_{m}^{t+\tau, \nu}) g^{\prime}(x_{k}^{t+\tau, \nu}) y_{n}^{t+\tau, \nu}$ and $\frac{1}{V} \sum_{\nu=1}^V g(x_{i}^{t+\tau, \nu}) g^{\prime}(x_{k}^{t+\tau, \nu}) y_{n}^{t+\tau, \nu}$ that appeared in Eq.~(\ref{eq:D_R_for_self_average}) converge to their average values $I_3(k, n, m)$ and $I_3(k, n, i)$.

We then further consider the limit where $\Delta t \to \infty$ but $\Delta t / N = \dd \alpha \to 0$. In this limit, $\alpha$ plays the role of a continuous time variable, and the term $\frac{1}{N} \sum_{\tau=1}^{\Delta t} (\cdots)$ in the right-hand side of Eq.~(\ref{eq:D_R_for_self_average}) involves the summation of large number of random variables, where the randomness comes from $\{ u_{m}^{t+\tau} \}$. We then invoke the law of large numbers and apply $\frac{1}{N} \sum_{\tau=1}^{\Delta t} (\cdots) \to \langle \cdots \rangle_{ \vec{u} }$ in Eq.~(\ref{eq:D_R_for_self_average}), which heuristically justifies the self-averaging property of dynamics of $R_{kn}$. 

Similar considerations apply to the dynamics of $Q_{kl}$, except that one needs to deal with averages of higher order terms like $\frac{1}{V} \sum_{\nu=1}^V g(y_{n}^{t+\tau, \nu})  g(y_{m}^{t+\tau, \nu}) g^{\prime}(x_{k}^{t+\tau, \nu}) g^{\prime}(x_{l}^{t+\tau, \nu})$, $\frac{1}{V} \sum_{\nu=1}^V g(y_{n}^{t+\tau, \nu})  g(x_{i}^{t+\tau, \nu}) g^{\prime}(x_{k}^{t+\tau, \nu}) g^{\prime}(x_{l}^{t+\tau, \nu})$ and $\frac{1}{V} \sum_{\nu=1}^V g(x_{i}^{t+\tau, \nu}) g(x_{j}^{t+\tau, \nu}) g^{\prime}(x_{k}^{t+\tau, \nu}) g^{\prime}(x_{l}^{t+\tau, \nu})$, which are usually subject to larger statistical fluctuations. As well, these terms also have prefactors that comprise of higher order terms of the learning rates like $\eta_J^2 \eta_w^2$, $\eta_J^2 \eta_w^3$ and $\eta_J^2 \eta_w^4$. Therefore, higher values of learning rates can magnify the approximation or statistical errors when $V$ is finite, as we have seen in Fig.~\ref{fig:E_D} .
In Fig.~\ref{fig:compare_Q_vary_etaw} and Fig.~\ref{fig:compare_Q_vary_V}, we show that for smaller values of $\eta_w$ and larger values of $V$, the theory provides a more accurate description of evolution of the order parameters $\{ Q_{kl} \}$.

\begin{figure}
    \centering
    \includegraphics[scale=1]{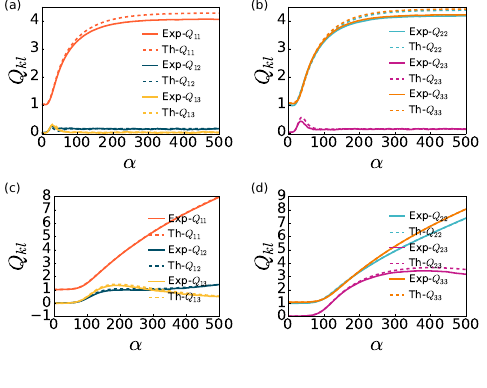}
    \caption{Comparison of the theoretical and numerical results of $\{ Q_{kl} \}$ for different values of $\eta_w$, where the experiment setup is the same as the one in Sec.~\ref{sec:compare}, except that a single run of the simulated dynamics is considered here. Solid lines represent experimental data and dashed lines correspond to theoretical predictions. 
    Panels (a) and (b) correspond to $\eta_w = 4$. Panels (c) and (d) correspond to $\eta_w = 1$.}
    \label{fig:compare_Q_vary_etaw}
\end{figure}

\begin{figure}
    \centering
    \includegraphics[scale=1.25]{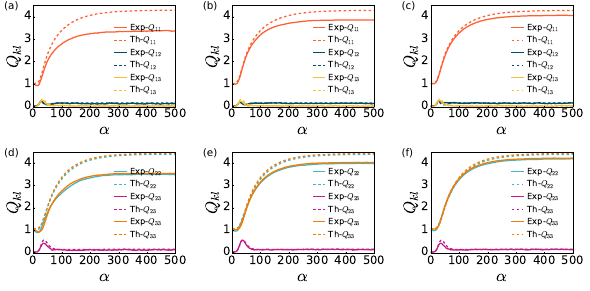}
    \caption{Comparison of the theoretical and numerical results of $\{ Q_{kl} \}$ for different values of $P,V$, where the experiment setup is the same as the one in Sec.~\ref{sec:compare}, except that a single run of the simulated dynamics is considered here.
    Solid lines represent experimental data and dashed lines correspond to theoretical predictions. 
    Panels (a) and (d) correspond to $P=V=20$. Panels (b) and (e) correspond to $P=V=50$. Panels (c) and (f) correspond to $P=V=100$.}
    \label{fig:compare_Q_vary_V}
\end{figure}

\section{Details on the Variability of Teacher Representation Mapping} \label{sec:details_dB}
In this appendix, we provide more details of the case where the teacher networks of different tasks have different representation mapping as introduced in Sec.~\ref{sec:dB}. In this case, the $t$th task produces samples as
\begin{align}
    \sigma^t = \frac{1}{\sqrt{M}} \sum_{m=1}^{M} u^t_m \; g\big[ (\gamma\vec{B}_m + \sqrt{1-\gamma^{2}}\Delta\vec{B}_m^t)^{\top} \cdot \vec{\xi} \big]. 
\end{align}
Assuming that $\gamma$ is close to $1$, we apply the Taylor series expansion to the above equation and keep the first-order terms, which results in
\begin{align}
    \sigma^t \approx \frac{1}{\sqrt{M}} \sum_{m=1}^{M} u_m^t \; \big[ g(\gamma\vec{B}_m^{\top} \cdot \vec{\xi}) + \sqrt{1-\gamma^{2}}g^{\prime}(\gamma\vec{B}_m^{\top} \cdot \vec{\xi}) (\Delta\vec{B}_m^{t \top} \cdot \vec{\xi}) \big]. 
\end{align}
The loss in the inner-loop optimization is computed as
\begin{align}
    \ell^{t}_{\text{inner}} = \frac{1}{2P}\sum_{\mu=1}^{P} \bigg [\sum_{m=1}^{M}\frac{u^{t}_{m}[g(\gamma y_{m}^{t,\mu}) + \sqrt{1-\gamma^{2}}g^{\prime}(\gamma y_{m}^{t,\mu})(\Delta\vec{B}^{t \top}_{m} \cdot \vec{\xi}^{t,\mu})]}{\sqrt{M}}-\sum_{k=1}^{K}\frac{w_{k}^{t}g(x_{k}^{t, \mu})}{\sqrt{K}} \bigg]^2.  
\end{align}
The hidden-to-output weight $w_k$ are then updated as (assuming that it starts from the initial value $w_k^0 = 0$)
\begin{align}
    w_{k}^{t} & = - \eta_{w}\frac{\partial \ell^{t}_{\text{inner}}}{\partial w_{k}} \bigg\vert_{w_k = 0} \nonumber \\
    & = \frac{\eta_{w}}{P}\sum_{\mu=1}^{P}\sum_{m=1}^{M}\frac{u^{t}_{m}[g(\gamma y_{m}^{t,\mu}) + \sqrt{1-\gamma^{2}}g^{\prime}(\gamma y_{m}^{t,\mu})(\Delta\vec{B}^{t,\top}_{m} \cdot \vec{\xi}^{t,\mu})]}{\sqrt{M}}\frac{g(x_{k}^{t, \mu})}{\sqrt{K}}. 
\end{align}

Similarly, the loss in the outer-loop optimization is
\begin{align}
    \ell_{\text{outer}}^{t} 
    & = \frac{1}{2V}\sum_{\nu=1}^{V} \bigg[ \sum_{m=1}^{M}\frac{u^{t}_{m}[g(\gamma y_{m}^{\n}) + \sqrt{1-\gamma^{2}}g^{\prime}(\gamma y_{m}^{\n})(\Delta\vec{B}^{t, \top}_{m} \cdot \vec{\xi}^{\n})]}{\sqrt{M}}-\sum_{k=1}^{K}\frac{ w^{t}_{k} g(x_{k}^{t, \nu})}{\sqrt{K}} \bigg]^2. 
\end{align}
Therefor, the parameters $\vec{J}_k$ are updated based on this outer loss:
\begin{align}
    \vec{J}_{k}^{t} 
    & = \vec{J}_{k}^{t-1} - \frac{\eta_{J}}{N}\frac{\partial \ell_{\text{outer}}^{t}}{\partial \vec{J}_{k}}\\
    &= \vec{J}_{k}^{t-1}+\frac{\eta_{J}}{NV}\sum_{\nu=1}^{V} \bigg\{\sum_{m=1}^{M}\frac{u^{t}_{m}[g(\gamma y_{m}^{\n}) + \sqrt{1-\gamma^{2}}g^{\prime}(\gamma y_{m}^{\n})(\Delta\vec{B}^{t \top}_{m} \cdot \vec{\xi}^{\n})]}{\sqrt{M}}-\sum_{i=1}^{K}\frac{ w^{t}_{i} g(x_{i}^{t, \nu})}{\sqrt{K}} \bigg\} \frac{ w_{k}^{t}  g^{\prime}(x^{t, \nu}_{k})\vec{\xi}^{t, \nu}}{\sqrt{K}}.  \nonumber
\end{align}

By defining $h_k^{\n}$ as
\begin{align}
   h_{k}^{\n} 
   &= \bigg\{ \sum_{m=1}^{M}\frac{u^{t}_{m}[g(\gamma y_{m}^{\n}) + \sqrt{1-\gamma^{2}}g^{\prime}(\gamma y_{m}^{(\n)})( \Delta\vec{B}^{t \top}_{m} \cdot \vec{\xi}^{\n})]}{\sqrt{M}}-\sum_{i=1}^{K}\frac{ w^{t}_{i} g(x_{i}^{t, \nu})}{\sqrt{K}} \bigg\} w_{k}^{t}  g^{\prime}(x^{t, \nu}_{k}),
\end{align}
the order parameters $R$, $Q$ are calculated as
\begin{align}
    R_{kn}^{t} - R_{kn}^{t-1} 
    =& \frac{\eta_{J}}{NV}\sum_{\nu=1}^{V}\frac{h_{k}^{\n}y_{n}^{\n}}{\sqrt{K}}, \\
    Q_{kl}^{t}-Q_{kl}^{t-1} = 
    & \frac{\eta_{J}}{NV}\sum_{\nu=1}^{V}\frac{h_{k}^{\n}x_{l}^{\n}}{\sqrt{K}} + \frac{\eta_{J}}{NV}\sum_{\nu=1}^{V}\frac{h_{l}^{\n}x_{k}^{\n}}{\sqrt{K}} + \frac{\eta_{J}^{2}}{NV^{2}}\sum_{\nu=1}^{V}\frac{h_{k}^{\n}h_{l}^{\n}}{K}.
\end{align}
By following the same procedures as those in the main text, we can obtain the dynamical equations of the order parameters $\{ R_{kn}, Q_{kl} \}$.

\section{L2-norm Regularization} \label{sec:L2reg}
In this appendix, we enrich the modeling of meta-learning by considering two additional elements: (i) the outputs are noisy and (ii) L2-norm regularization for the representation mapping $\mat{J}$ is used. For the latter case, it can help to stabilize the solution of the meta-learner's weights $\mat{J}$. 

The teacher output is perturbed by some noise as
\begin{align}
    \sigma^{\m} & = \frac{1}{\sqrt{M}}\sum_{m=1}^{M}u_{m}^{t}g(y_{m}^{\m}) + \Delta^{\m}, \quad \Delta^{\m} \sim \mathcal{N}(0, \Sigma_{\text{noise}}).
\end{align}

The loss in the inner loop optimization is 
\begin{align}
    \ell_{\text{inner}}^{t} & = \frac{1}{2P}\sum_{\mu =1}^{P} \bigg[ \frac{1}{\sqrt{M}}\sum_{m=1}^{M}u_{m}^{t}g(y_{m}^{\m}) + \Delta^{\m} - \frac{1}{\sqrt{K}}\sum_{k=1}^{K}w_{k}g(x_{k}^{\m}) \bigg]^{2}.
\end{align}

The loss in the outer loop optimization is
\begin{align}
    \ell_{\text{outer}}^{t} & = \frac{1}{2V}\sum_{\nu =1}^{V}\bigg[\frac{1}{\sqrt{M}}\sum_{n=1}^{M}u_{n}^{t}g(y_{n}^{\n}) + \Delta^{\n} - \frac{1}{\sqrt{K}}\sum_{i=1}^{K}w_{i}g(x_{i}^{\n})\bigg]^{2} + \frac{\lambda}{2} \big|\big| \mat{J} \big|\big |^{2}.
\end{align}

So we can update the parameter $\vec{J}_{k}$ as
\begin{align}
    \vec{J}_{k}^{t} & = \big(1-\frac{\lambda\eta_J}{N}\big) \vec{J}_{k}^{t-1} \nonumber \\
    & + \frac{\eta_{J}}{N\sqrt{K}} \frac{1}{V}\sum_{\nu = 1}^{V} \bigg[ \frac{1}{\sqrt{M}}\sum_{n=1}^{M}u_{n}^{\n}g(\vec{B}_{n} \cdot \vec{\xi}^{\n}) + \Delta_{\text{noise}}^{\n} - \frac{1}{\sqrt{K}}\sum_{i=1}^{K}  w_{i}^{\n} g(\vec{J}_{i}\cdot \vec{\xi}^{\n}) \bigg]  w_{k}^{\n}  g^\prime(\vec{J}_{k}\cdot \vec{\xi}^{\n}) \; \vec{\xi}^{\n}. 
\end{align}

Following the same procedures as those in the main text, the dynamical equations of the order parameters are
\begin{align}
    \frac{\dd R_{kn}}{\dd \alpha} = &\frac{\eta_{J}\eta_{w}}{KM}\sum_{m=1}^{M}f_{km}I_3(k, n, m) -\frac {\eta_{J}\eta_{w}^{2}}{K^{2}M}\sum_{i=1}^{K}\sum_{m=1}^{M}f_{im}f_{km}I_3(k, n, i) - \lambda\eta_J R_{kn}, \\
    \frac{\dd Q_{kl}}{\dd \alpha} = & \frac{\eta_{J}\eta_{w}}{KM}\sum_{n=1}^{M}[f_{ln}I_3(l, k, n)+ f_{kn}I_3(k, l, n)] \nonumber \\
    & - \frac{\eta_{J}\eta_{w}^{2}}{K^{2}M}\sum_{i=1}^{K}\sum_{n=1}^{M}[f_{in}f_{ln}I_3(l, k, i) + f_{in}f_{kn}I_3(k, l, i)] \nonumber \\
    & + \frac{(\eta_{J}\eta_{w})^{2}}{V(KM)^{2}} \bigg[ \sum_{n=1}^{M}\sum_{m=1}^{M} \big[ f_{km}f_{lm}I_4(k, l, n, n) + (f_{kn}f_{lm} + f_{km}f_{ln})I_4(k,l,n,m) \big] + \sum_{n=1}^{M}f_{kn}f_{ln}\Sigma_{\text{noise}}I_2^{\prime}(k, l) \bigg] \nonumber \\
    & -\frac{2\eta_{J}^{2}\eta_{w}^{3}}{VK(KM)^{2}}\sum_{j=1}^{K}\sum_{n=1}^{M}\sum_{m=1}^{M}[f_{jn}f_{km}f_{lm} + f_{jm}f_{kn}f_{lm} + f_{jm}f_{km}f_{ln}]I_4(k,l,n,j) \nonumber \nonumber \\
    & + \frac{\eta_{J}^{2}\eta_{w}^{4}}{VK^{2}(KM)^{2}}\sum_{i=1}^{K}\sum_{j=1}^{K}\sum_{n=1}^{M}\sum_{m=1}^{M}[f_{in}f_{jn}f_{km}f_{lm} +f_{in}f_{jm}f_{kn}f_{lm} + f_{in}f_{jm}f_{km}f_{ln}]I_4(k, l, i, j) \nonumber \\
    & - 2\lambda\eta_J Q_{kl}, 
\end{align}

where $I_2^{\prime}(k,l)$ can be calculated as
\begin{align}
    I_2^{\prime}(k,l) & =\big \langle g^{\prime}(x_k)g^{\prime}(x_l)\big \rangle \nonumber\\
    & = \frac{2}{\pi}\frac{1}{\sqrt{\big| \mat{I} + \mat{\Sigma}_{kl}\big|}},
\end{align}
with
\begin{align}
    \mat{\Sigma}_{kl} & := \left(\begin{matrix}Q_{kk} &  Q_{kl} \\ Q_{kl} & Q_{ll} \nonumber\\
    \end{matrix}\right). 
\end{align}

Based on $R$ and $Q$, we can calculate the meta-generalization error as
\begin{align}
    \epsilon_{g}^{\text{meta}} = & \frac{1}{2M}\bigg[\sum_{n=1}^{M}\arcsin \frac{T_{nn}}{\sqrt{1 + T_{nn}}\sqrt{1 + T_{nn}}} + \Sigma_{\text{noise}} \bigg] - \frac{\eta_{w}}{KM} \sum_{k=1}^{K}\sum_{n=1}^{M}\arcsin^{2} \big( \frac{R_{kn}}{\sqrt{1 + Q_{kk} }\sqrt{1 + T_{nn}}} \big) \\
    & + \frac{1}{2}\frac{\eta_{w}^{2}}{K^{2}M}\sum_{k,l =1}^{K}\sum_{n=1}^{M}\arcsin \big( \frac{Q_{kl}}{\sqrt{1 + Q_{kk}}\sqrt{1 + Q_{ll}}} \big) \arcsin \big(\frac{R_{kn}}{\sqrt{1 + Q_{kk}}\sqrt{1 + T_{nn}}} \big) \arcsin \big( \frac{R_{ln}}{\sqrt{1 + Q_{ll}}\sqrt{1 + T_{nn}}} \big). \nonumber
\end{align}

\section{Linear Activation Function} \label{sec:linear_act_func}
\subsection{Dynamical equations}
Throughout the paper, we have analyzed meta-learning on networks with nonlinear activation functions. Here we turn our attention to two-layer networks with a linear activation function $g(x) = x$ as considered in~\cite{Collins2022, Yuksel2024}.
In this case, it is possible for each hidden unit of the meta-learner to learn a superposition of all hidden units of the meta-teachers, so we need to examine the overlap order parameters more carefully to determine weather the teacher's representation mappings $\mat{B}$ have been properly learned.

The expected hidden-to-output weights for a specific task $\mathcal{T}_{t}$ can be expressed as
\begin{align}
    \langle w_{k}^{t}\rangle_{\{\vec{\xi}^{\m}\}} & =\frac{\eta_w}{P} \bigg\langle \sum_{\mu =1}^{P} \sigma^{\m} \frac{[(\vec{J}_k^{t})^{T}\cdot \vec{\xi}^{\m}]}{\sqrt{K}} \bigg\rangle_{\{\vec{\xi}^{\m}\}} \nonumber \\
    &  = \frac{\eta_w
    }{\sqrt{KM}}\sum_{n=1}^{M}u_{n}R_{kn}. 
\end{align}
Upon averaging over the training set $D_{\text{train}}^{\mathcal{T}_{t}}$, the updated input-to-hidden weights $\vec{J}_k$ can be expressed as
\begin{align}
    \vec{J}_{k}^{t} & =\vec{J}_{k}^{t-1}  + \frac{\eta_J}{VN}\sum_{\nu =1}^{V}\bigg[\frac{1}{\sqrt{M}}\sum_{n=1}^{M}u_{n}^{t}y_{n}^{\n}-\frac{1}{\sqrt{K}}\sum_{i=1}^{K}\langle w_{i}^{t}\rangle x_{i}^{\n}\bigg]\frac{\langle w_{k}^{t} \rangle \vec{\xi}^{\n}}{\sqrt{K}}, 
\end{align}
where $y_n^{\n} = \vec{B}_n^{\top} \cdot \vec{\xi}^{\n}$, $x_i^{\n} = \vec{J}_i^{\top} \cdot \vec{\xi}^{\n}$ and we have made use of the approximation $\langle w_i w_k \rangle_{ \{ \vec{\xi}^{\mu} \} } \approx \langle w_i \rangle_{ \{ \vec{\xi}^{\mu} \} } \langle w_k \rangle_{ \{ \vec{\xi}^{\mu} \} }$ as before.

Following the same procedures outlined in the main text, we obtain the following dynamical equations for the order parameters and the expression for the meta-generalization errors:
\begin{align}
    \frac{\dd R_{kn}}{\dd \alpha} & =  \frac{\eta_J\eta_w}{KM}\sum_{m=1}^{M}T_{mn}R_{km}-\frac{\eta_J \eta_{w}^{2}}{K^{2}M}\sum_{i=1}^{K}\sum_{m=1}^{M}R_{in}R_{im}R_{km}, \\
     \frac{\dd Q_{kl}}{\dd \alpha} &=    \frac{\eta_J\eta_w}{KM}\sum_{n=1}^{M}R_{ln}R_{kn}-\frac{\eta_J \eta_{w}^{2}}{K^{2}M}\sum_{i=1}^{K}\sum_{n=1}^{M}Q_{il}R_{in}R_{kn} \nonumber \\
    & \quad + \frac{\eta_J\eta_w}{KM}\sum_{n=1}^{M}R_{ln}R_{kn}-\frac{\eta_J \eta_{w}^{2}}{K^{2}M}\sum_{i=1}^{K}\sum_{n=1}^{M}Q_{ik}R_{in}R_{ln} \nonumber\\
    & \quad + \frac{\eta_J^{2}}{V}\frac{\eta_w^{2}}{K^{2}M^{2}}\sum_{n=1}^{M}\sum_{m=1}^{M}\big[T_{nn}R_{km}R_{lm} + T_{nm}R_{kn}R_{km} + T_{nm}R_{km}R_{ln} \big] \nonumber \\
    & \quad - \frac{2\eta_{J}^{2}}{V}\frac{\eta_w^3}{K^{3}M^{2}}\sum_{i=1}^{K}\sum_{n=1}^{M}\sum_{m=1}^{M}R_{in}\big[R_{in}R_{km}R_{lm} + R_{im}R_{kn}R_{km} + R_{im}R_{km}R_{ln} \big] \nonumber  \\
    & \quad + \frac{\eta_{J}^{2}}{V}\frac{\eta_w^4}{K^{4}M^{2}}\sum_{i=1}^{K}\sum_{j=1}^{K}\sum_{n=1}^{M}\sum_{m=1}^{M}Q_{ij}\big[R_{in}R_{jn}R_{km}R_{lm} + R_{in}R_{jm}R_{kn}R_{km} + R_{in}R_{jm}R_{km}R_{ln} \big], \\
    \epsilon_{g}^{\text{meta}} & = \frac{1}{2M}\sum_{n=1}^{M}T_{nn} -\frac{\eta_w}{KM}\sum_{n=1}^{M}\sum_{k=1}^{K}R_{kn}^{2} + \frac{\eta_w^2}{2K^{2}M}\sum_{k=1}^{K}\sum_{l=1}^{K}\sum_{n=1}^{M}Q_{kl}R_{kn}R_{ln}.  
\end{align}

\subsection{Case study} \label{sec:lin_act_func_case}
We consider a case study where the system parameters are set as $K = M = 3, \eta_J = 3, \eta_w = 0.5, V=100, T_{mn} = m\delta_{m,n}$, and the initial condition for the order parameters $Q$ as $Q_{kl} = \frac{1}{2}\delta_{kl}$. We investigate three different choices of initial conditions for the order parameters $R$:

(i) In the first condition, $R_{kn} = 1 \times 10^{-12}$ at $\alpha = 0$.

(ii) In the second condition,  $R_{11} = 1.1 \times 10^{-12}$ with the remaining elements set as $ R_{kn} = 1 \times 10^{-12}$ at $\alpha = 0$.

(iii) In the third condition, $R_{11} = 1.1 \times 10^{-12} $, $ R_{31} = 1.2 \times 10^{-12}$, with the remaining elements set as $R_{kl} = 1 \times 10^{-12}$ at $\alpha = 0$.

These scenarios yield three distinct types of learning behaviors as shown in Fig.~\ref{fig:000}, Fig.~\ref{fig:1_13_1.1e-12_0.1}, and Fig.~\ref{fig:1_13_1.1e-12_0.1_0.2}.
By examining the meta-generalization error in these figures, it can be observed that $\epsilon_g^{\text{meta}}$ becomes trapped in suboptimal solutions under the first and second conditions up to $\alpha  = 500$, while it decreases to $0$ in the third condition.

To gain further insight, we examine the evolution of $Q$ and $\rho$. In the second choice of initial conditions, the results in Fig.~\ref{fig:000}(b) show that $\{ Q_{kk} \}$ converges to the same value, and $\{ Q_{kl} \mid k \neq l \}$ converge another value. This indicates that the input-to-hidden weights $\vec{J}_1$, $\vec{J}_2$, and $\vec{J}_3$ have an equal length but they are not identical. Additionally, Fig.~\ref{fig:000}(d) shows that $0 < \rho_{13} = \rho_{23} = \rho_{33} < 1$, while $\rho_{kn} = 0$ for all other $k$ and $n$. This suggests that $\vec{J}_{1}$, $\vec{J}_{2}$, and $\vec{J}_{3}$ have overlap with $\vec{B}_3$. Furthermore, they are orthogonal to $\vec{B}_{1}$ and $\vec{B}_{2}$. Thus, for the meta-teacher's linear representation space spanned by $\{ \vec{B}_{1}, \vec{B}_{2}, \vec{B}_{3} \}$, only one direction along $\vec{B}_3$ has been learned by the meta-learner.

In the second choice of initial conditions, the results in Fig.~\ref{fig:1_13_1.1e-12_0.1}(d) show that $\rho_{11} = \rho_{21} = \rho_{31} = 0$, while the other $\rho_{kn}$ values are nonzero and differ from $1$. This implies that $\vec{J}_1$, $\vec{J}_2$, and $\vec{J}_3$ have overlaps with $\vec{B}_2$ and $\vec{B}_3$ but remain orthogonal to $\vec{B}_1$. As shown in Fig.~\ref{fig:1_13_1.1e-12_0.1}(b), $Q_{22} = Q_{33} \neq Q_{11}$, $Q_{12} = Q_{13} = 0$, and $Q_{23} \neq 0$, indicating that $\vec{J}_1$ is orthogonal to $\vec{J}_2$ and $\vec{J}_3$. Therefore, in this case, the meta-learner has learned a two-dimensional subspace of the meta-teacher's representation mapping.

Finally, in the third choice of initial conditions, the results in Figure~\ref{fig:1_13_1.1e-12_0.1_0.2}(d) show that all $\rho_{kn}$ values are non-zero and distinct from one. Figure~\ref{fig:1_13_1.1e-12_0.1_0.2}(b) shows that $Q_{11} = Q_{22} = Q_{33} \neq 0$ and $Q_{kl} = 0$ for $k \neq l$, indicating that $\vec{J}_1$, $\vec{J}_2$, and $\vec{J}_3$ are mutually orthogonal. We can deduce that the linear space spanned by $\{ \vec{J}_1, \vec{J}_2, \vec{J}_3 \}$ coincides with the meta-teacher network's representation space spanned by $\{ \vec{B}_1, \vec{B}_2, \vec{B}_3 \}$. Therefore, in this case, the student network has fully learned the teacher network's representation, and as a result, the meta-generalization error is able to decrease to zero as shown in Fig.~\ref{fig:1_13_1.1e-12_0.1_0.2}(c).
These results are consistent with the findings in Refs.~\cite{Collins2022} and \cite{Yuksel2024}. 
The results also suggest that a higher degree of heterogeneity in the initial conditions, especially for $\{ R_{kn} \}$, is needed for effective meta-learning in linear networks.

\begin{figure}[!htbp]
    \centering
    \includegraphics[scale=1]{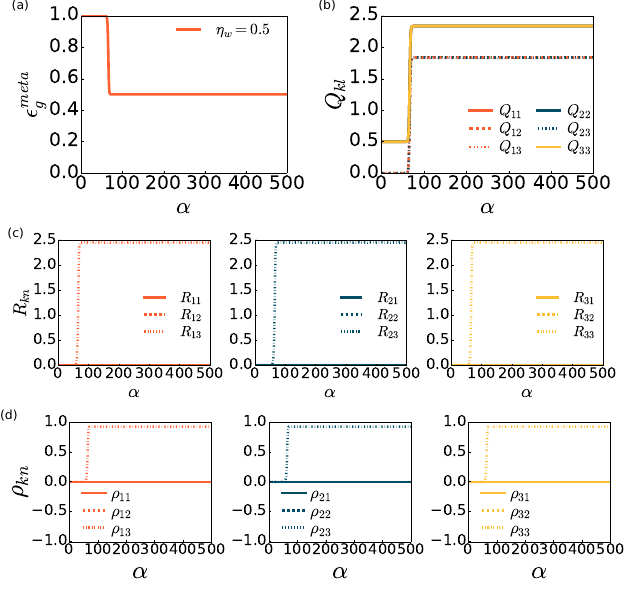} 
    \caption{Evolution of the meta-generalization error and the quantities $Q$, $R$, and $\rho$ in the case with linear activation function, with the system parameters specified in Sec.~\ref{sec:lin_act_func_case}.
    The first set of initial condition for $\{ R_{kn} \}$ is used, i.e., $R_{kn} = 1 \times 10^{-12}$ at $\alpha = 0$. 
    (a) The evolution of the meta-generalization error. (b) $Q_{kl}$ vs $\alpha$. (c) $R_{kn}$ vs $\alpha$. (d) $\rho_{kn}$ vs $\alpha$.}
    \label{fig:000}
\end{figure}

\begin{figure}[!htbp]
    \centering
    \includegraphics[scale=1]{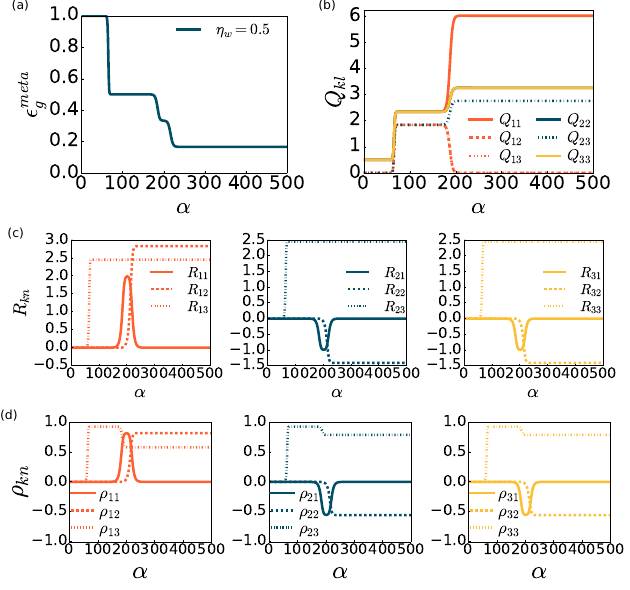} 
    \caption{Evolution of the meta-generalization error and the quantities $Q$, $R$, and $\rho$ in the case with linear activation function, with the system parameters specified in Sec.~\ref{sec:lin_act_func_case}.
    The second set of initial condition for $\{ R_{kn} \}$ is used, i.e., $R_{11} = 1.1 \times 10^{-12}$ with the remaining elements set as $ R_{kn} = 1 \times 10^{-12}$ at $\alpha = 0$. 
    (a) The evolution of the meta-generalization error. (b) $Q_{kl}$ vs $\alpha$. (c) $R_{kn}$ vs $\alpha$. (d) $\rho_{kn}$ vs $\alpha$.}
    \label{fig:1_13_1.1e-12_0.1}
\end{figure}

\begin{figure}[!htbp]
    \centering
    \includegraphics[scale=1]{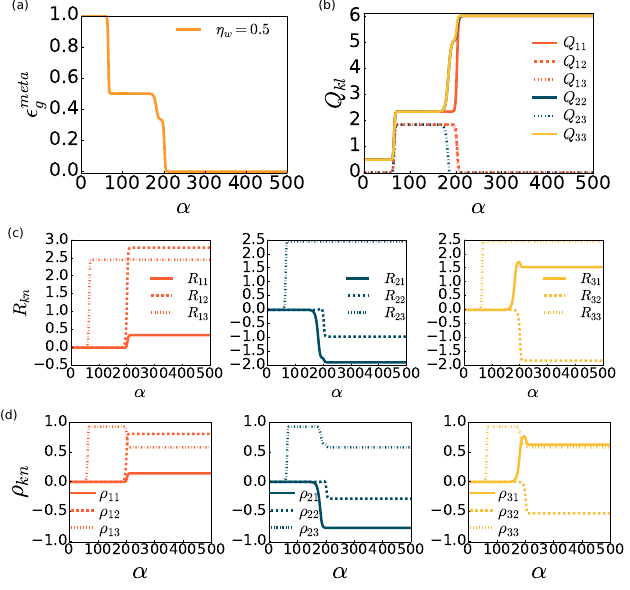} 
    \caption{Evolution of the meta-generalization error and the quantities $Q$, $R$, and $\rho$ in the case with linear activation function, with the system parameters specified in Sec.~\ref{sec:lin_act_func_case}.
    The third set of initial condition for $\{ R_{kn} \}$ is used, i.e., $R_{11} = 1.1 \times 10^{-12} $, $ R_{31} = 1.2 \times 10^{-12}$, with the remaining elements set as $R_{kl} = 1 \times 10^{-12}$ at $\alpha = 0$. 
    (a) The evolution of the meta-generalization error. (b) $Q_{kl}$ vs $\alpha$. (c) $R_{kn}$ vs $\alpha$. (d) $\rho_{kn}$ vs $\alpha$.}
    \label{fig:1_13_1.1e-12_0.1_0.2}
\end{figure}

\end{widetext}

\end{document}